\documentclass[sigconf]{aamas} 

\usepackage{balance} 

\usepackage{hyperref}

\usepackage{graphicx}

\usepackage{subfigure}

\usepackage{amsmath}
\usepackage{amsfonts}

\usepackage{amssymb}
\usepackage{textcomp}
\usepackage{xcolor}
\usepackage{multirow}
\usepackage{booktabs}

\usepackage{url}
\usepackage{newfloat}
\usepackage{listings}
\usepackage{algorithm}
\usepackage{algorithmic}

\setcopyright{ifaamas}
\acmConference[AAMAS '23]{Proc.\@ of the 22nd International Conference
on Autonomous Agents and Multiagent Systems (AAMAS 2023)}{May 29 -- June 2, 2023}
{London, United Kingdom}{A.~Ricci, W.~Yeoh, N.~Agmon, B.~An (eds.)}
\copyrightyear{2023}
\acmYear{2023}
\acmDOI{0}
\acmPrice{0}
\acmISBN{0}

\acmSubmissionID{180}

\title[AAMAS-2023 Formatting Instructions]{Infomaxformer: Maximum Entropy Transformer for Long Time-Series Forecasting Problem}

\author{Peiwang Tang}
\affiliation{
  \institution{Institute of Advanced Technology, University of Science and Technology of China}
  \city{Hefei 230026}
  \country{China}}
  \affiliation{
  \institution{G60 STI Valley Industry \& Innovation Institute, Jiaxing University}
  \city{Jiaxing 314001}
  \country{China}}
\email{tpw@mail.ustc.edu.cn}

\author{Xianchao Zhang}\authornote{corresponding author}
\affiliation{
  \institution{Key Laboratory of Medical Electronics and Digital Health of Zhejiang Province, Jiaxing University}
  \city{Jiaxing 314001}
  \country{China}}
  \affiliation{
  \institution{Engineering Research Center of Intelligent Human Health Situation Awareness of Zhejiang Provincey, Jiaxing University}
  \city{Jiaxing 314001}
  \country{China}}
\email{zhangxianchao@zjxu.edu.cn}

\newcommand{\BibTeX}{\rm B\kern-.05em{\sc i\kern-.025em b}\kern-.08em\TeX}

\begin{abstract}
The Transformer architecture yields state-of-the-art results in many tasks such as natural language processing (NLP) and computer vision (CV), since the ability to efficiently capture the precise long-range dependency coupling between input sequences. With this advanced capability, however, the quadratic time complexity and high memory usage prevents the Transformer from dealing with long time-series forecasting problem (LTFP). To address these difficulties: (i) we revisit the learned attention patterns of the vanilla self-attention, redesigned the calculation method of self-attention based the \textbf{Maximum Entropy Principle}. (ii) we propose a new method to sparse the self-attention, which can prevent the loss of more important self-attention scores due to random sampling.(iii) We propose Keys/Values Distilling method motivated that a large amount of feature in the original self-attention map is redundant, which can further reduce the time and spatial complexity and make it possible to input longer time-series. Finally, we propose a method that combines the encoder-decoder architecture with seasonal-trend decomposition, i.e., using the encoder-decoder architecture to capture more specific seasonal parts. A large number of experiments on several large-scale datasets show that our Infomaxformer is obviously superior to the existing methods. We expect this to open up a new solution for Transformer to solve LTFP, and exploring the ability of the Transformer architecture to capture much longer temporal dependencies.
\end{abstract}

\keywords{Maximum Entropy, Transformer, Time-Series, Forecasting}

\begin{document}
	
\pagestyle{fancy}
\fancyhead{}

\maketitle 



\section{Introduction}
Defined as an ordered dataset formed with time change, time-series refer to a series of ordered observations acquired according to time sequence \cite{cryer1986time}, which is widely used in commercial and industrial fields, such as biomedical field \cite{liu2018learning}, economic and financial field \cite{di2016artificial}, electric power \cite{zhou2021informer} and transportation field \cite{yin2016multivariate}. As an important part of time-series analysis, time-series forecasting mainly analyze the trend, periodicity, volatility and other time-series patterns of time-series by using the time-series data observed in history and the relevant rules that have been mastered, so as to predict the situation in the future \cite{lai2018modeling, bai2018empirical, wu2021autoformer}. In practical applications, we can use a large number of past time-series to achieve long-term prediction for the future, i.e., long time-series forecasting problem (LTFP). Recent deep prediction models have made great progress, especially Transformer based models \cite{jiang2022accurate, lim2021temporal, liu2021pyraformer, wu2020adversarial, tang2022mtsmae}. The Transformer \cite{vaswani2017attention} shows better performance than the recurrent neural network (RNN) model in modeling the long-term dependence of sequence data, and has achieved the best results in the natural language processing (NLP) \cite{devlin2018bert, radford2018improving} and computer vision (CV) \cite{2021An, he2022masked} fields, since its advanced self-attention mechanism.

However, there are still some problems in solving LTFP of existing Transformer models. First, the self-attention mechanism has high performance, but also brings high time complexity and memory usage \cite{liu2021swin, yu2021glance}. Although some large-scale Transformer models have produced impressive results in the NLP and CV fields \cite{radford2019language, bao2021beit}, they often require dozens or even hundreds of GPUs during training, which limits the possibility of Transformer models to solve LTFP. Although there have been some researches on reducing the time complexity and memory usage of the self-attention mechanism, only realize a limited reduce of complexity to $\mathcal{O}(L log L) $ \cite{zhou2021informer, li2019enhancing, kitaev2019reformer}. Moreover, some methods for reducing the complexity only randomly select dot-product pairs, which will cause some performance loss and lead to the long-term dependence of the sequences that cannot be well captured by the self-attention mechanism.
Second, it is unreliable to find the time dependence directly from the time-series, because these dependencies may be masked by the entangled temporal patterns.

\begin{figure*}[ht]
	\centering
	\includegraphics[scale=0.9]{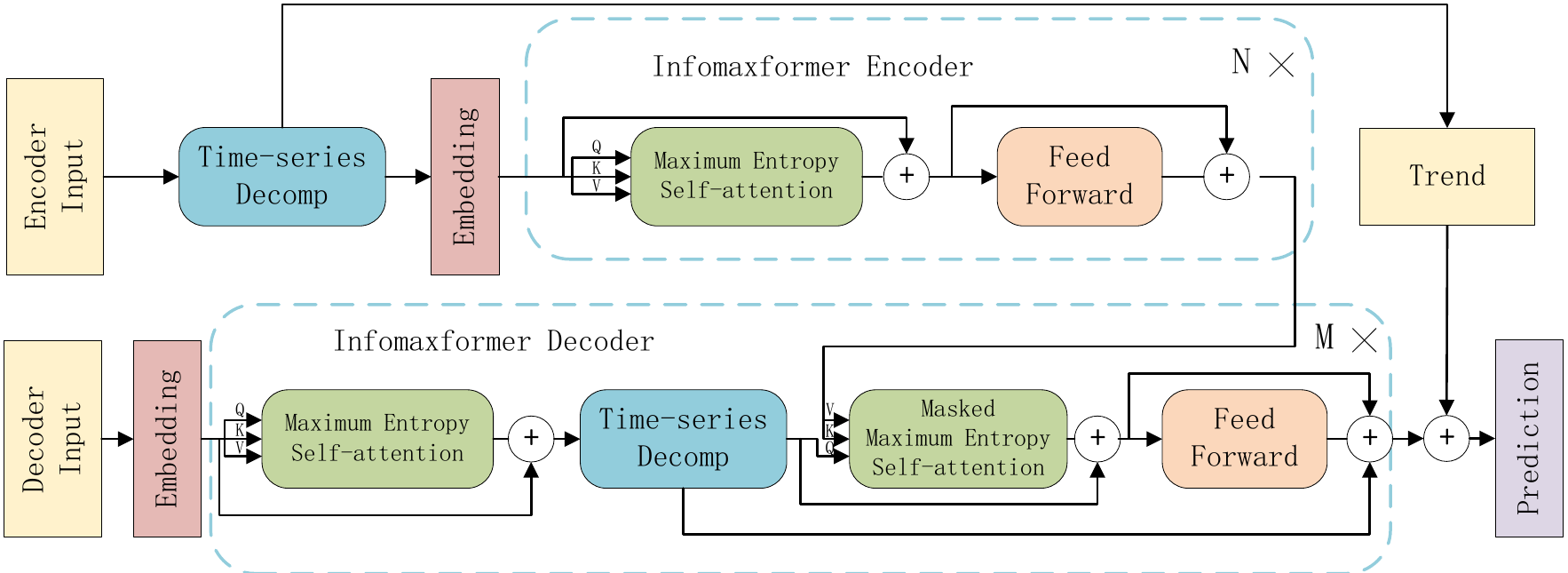}
	\caption{Infomaxformer architecture}
	\label{imgforemr}
\end{figure*}

In order to better solve LTFP, our work explicitly and deeply discussed the above problems, studied the sparsity of self-attention mechanism, decomposed the time-series, and updated the network components. Finally, we have conducted extensive experiments on five different datasets. The final experimental results show that our proposed Infomaxformer can significantly improve the accuracy of prediction, and is superior to other state-of-the-art models. The contributions of this paper are summarized as follows: 
\begin{itemize}
	\item We review the calculation method of self-attention mechanism from the perspective of information entropy \cite{shannon1948mathematical}, and sparse the calculation of self-attention by using the \textbf{Maximum Entropy Principle} \cite{jaynes1957information} to reduce the time complexity.
	\item In view of the data characteristic that local information of time-series is heavy spatial redundancy, we propose the Keys/Values Distilling method, which can further reduce the time and space complexity to $\mathcal{O}(L)$, and help the model to accept longer sequence inputs.
	\item In order to decompose time-series and explain complex time-series patterns, we propose a decomposition method, which is combined with self-attention mechanism, to process complex time-series and extract more useful features.
	\item We have conducted extensive experiments on datasets in many different fields, and the final results show that our proposed model achieves the most advanced performance in a variety of experimental settings.
\end{itemize}

\begin{figure*}[htbp]
	\centering
	\subfigure[Softmax scores at Head1@Encoder layer]{
		\begin{minipage}[t]{0.15\linewidth} 
			\includegraphics[width=\linewidth]{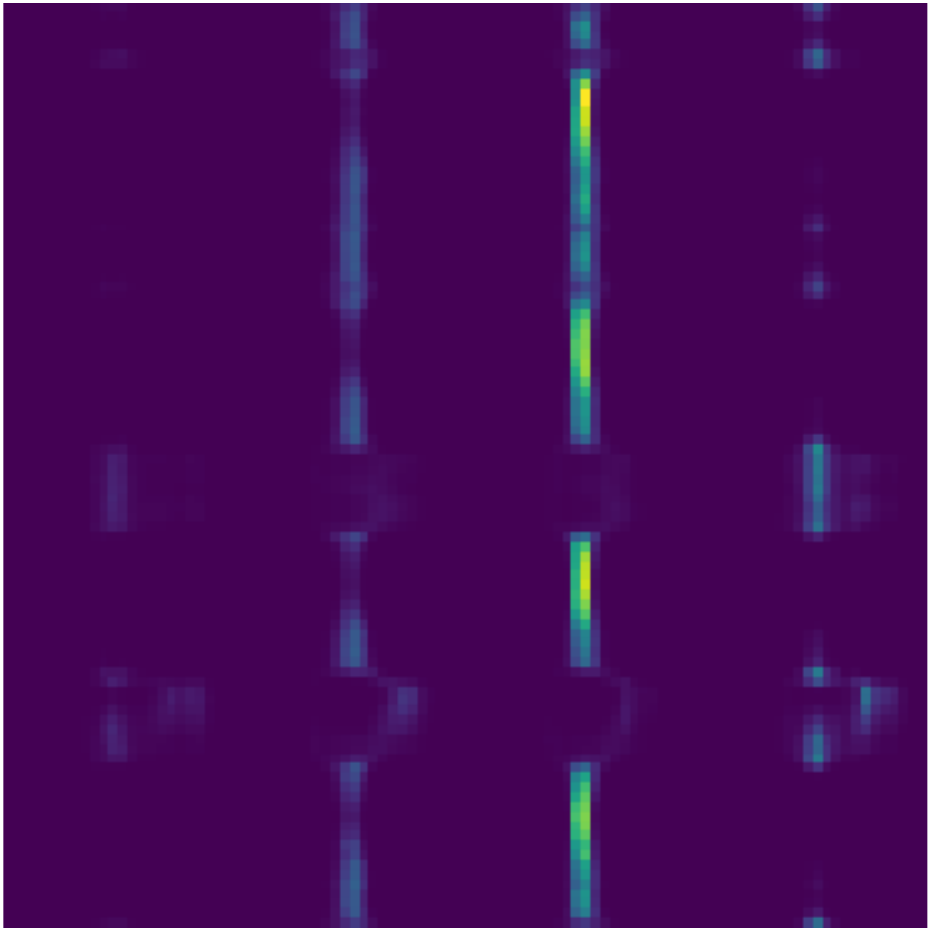}\vspace{1pt} 
			\includegraphics[width=\linewidth]{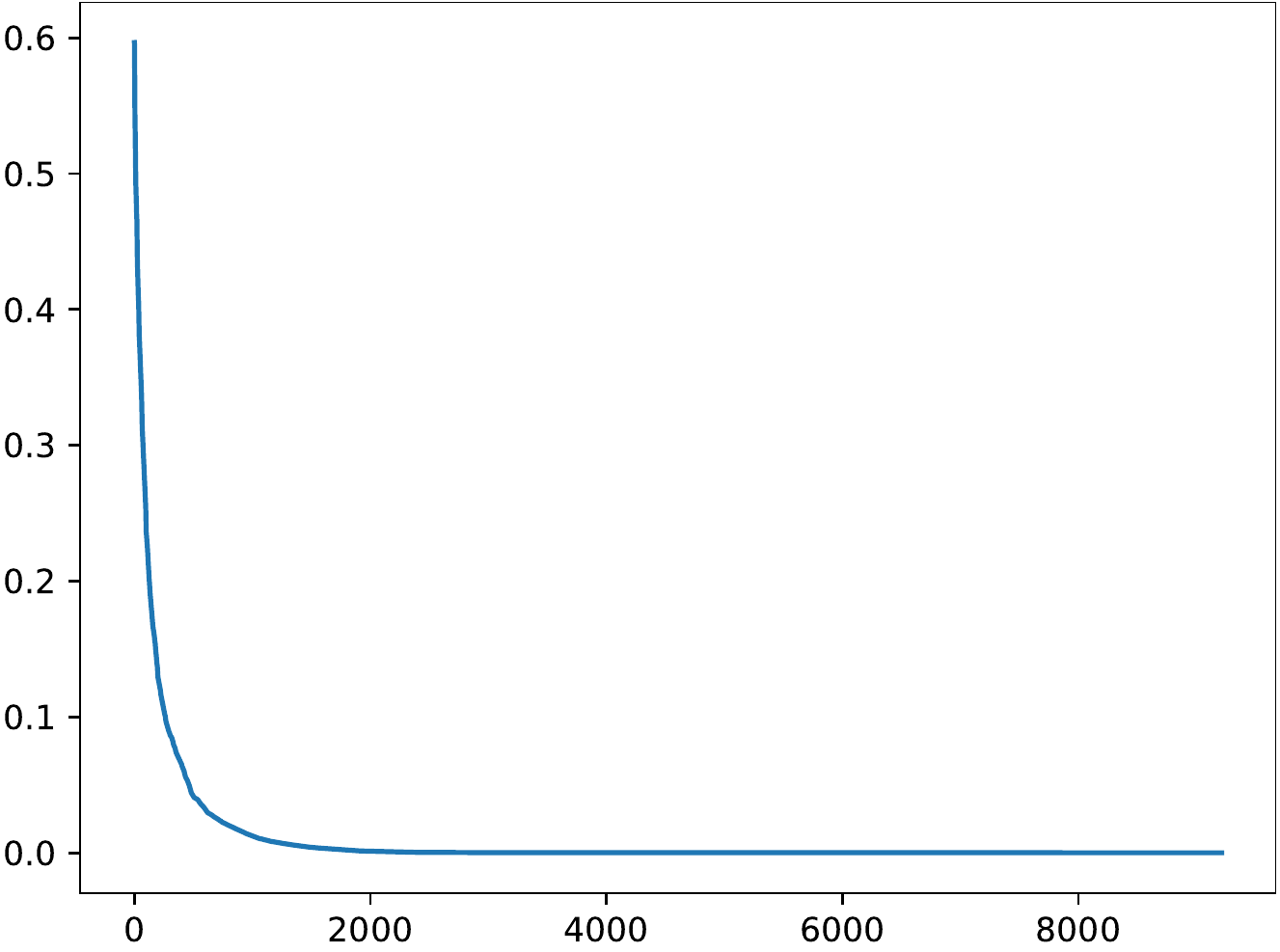}\vspace{1pt} 
		\end{minipage}
		\begin{minipage}[t]{0.15\linewidth} 
			\includegraphics[width=\linewidth]{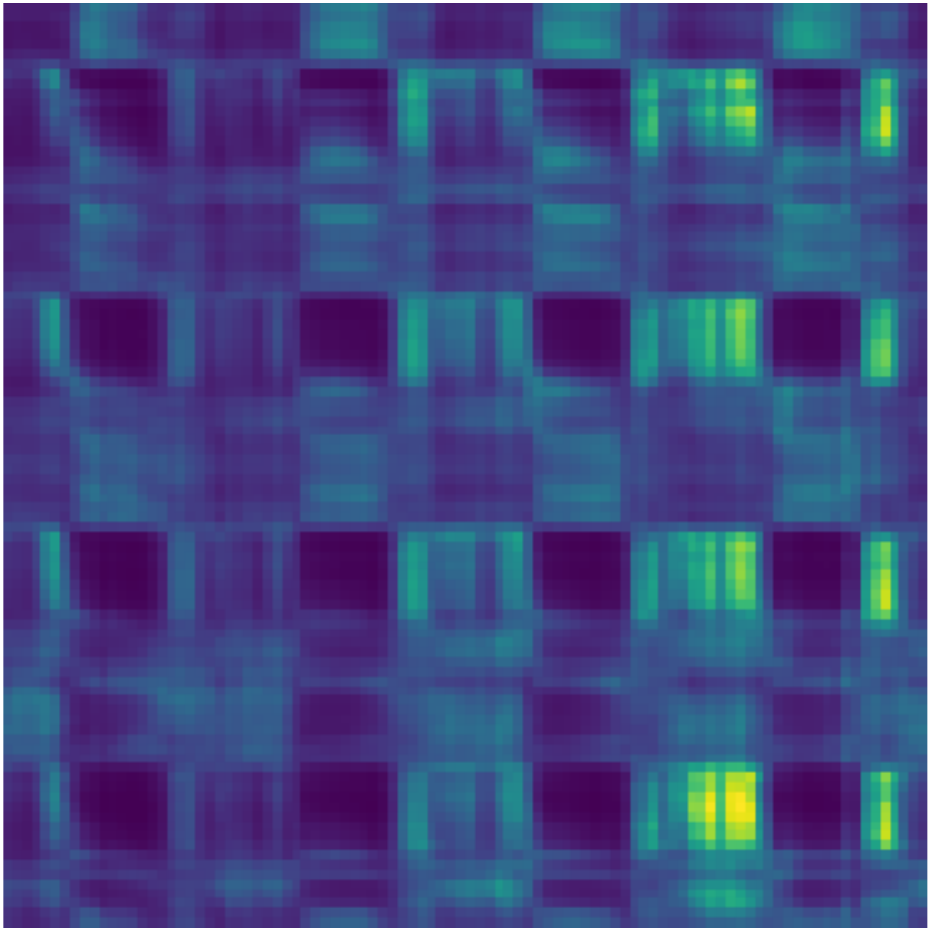}\vspace{1pt} 
			\includegraphics[width=\linewidth]{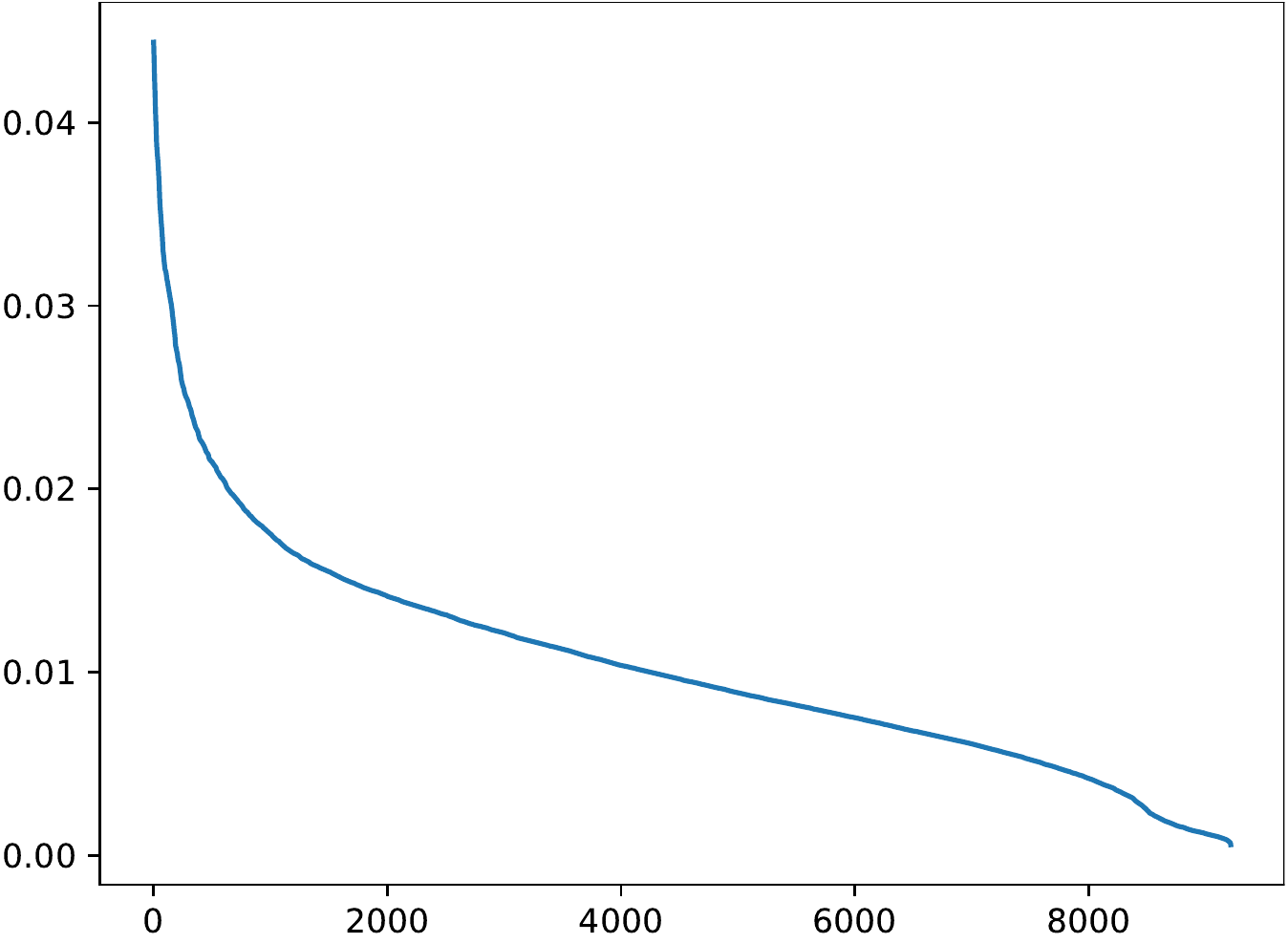}\vspace{1pt} 
		\end{minipage}
		\begin{minipage}[t]{0.15\linewidth} 
			\includegraphics[width=\linewidth]{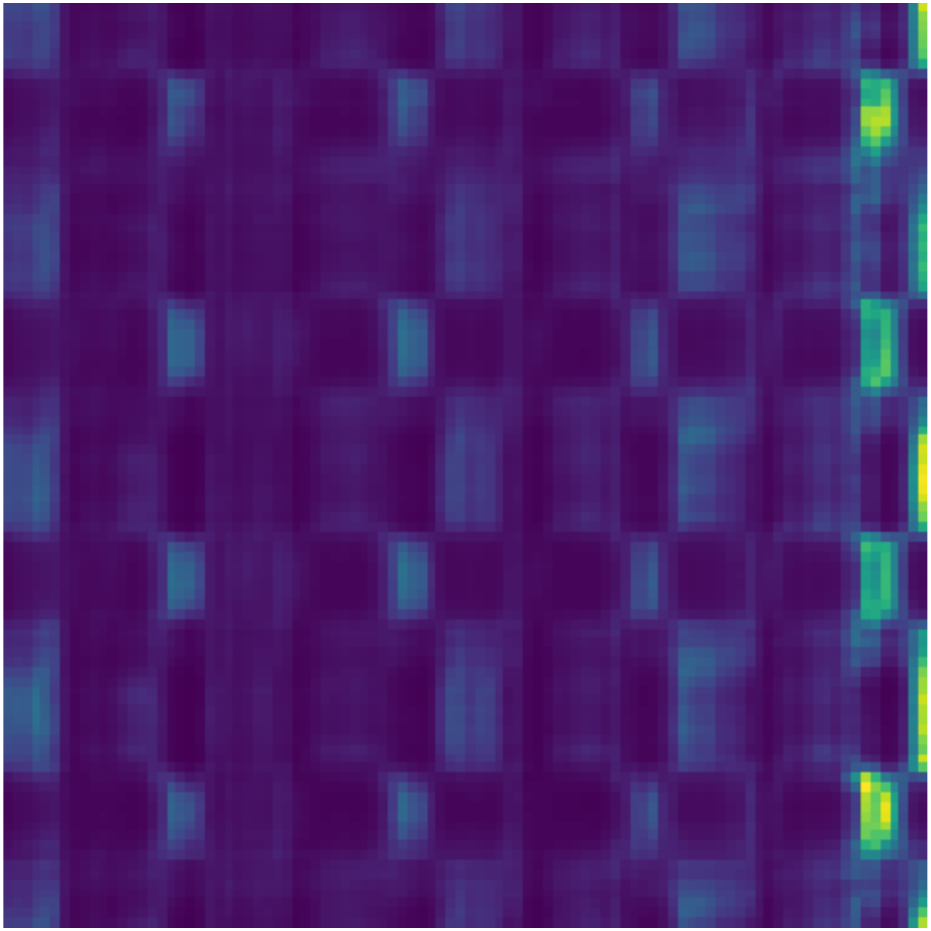}\vspace{1pt} 
			\includegraphics[width=\linewidth]{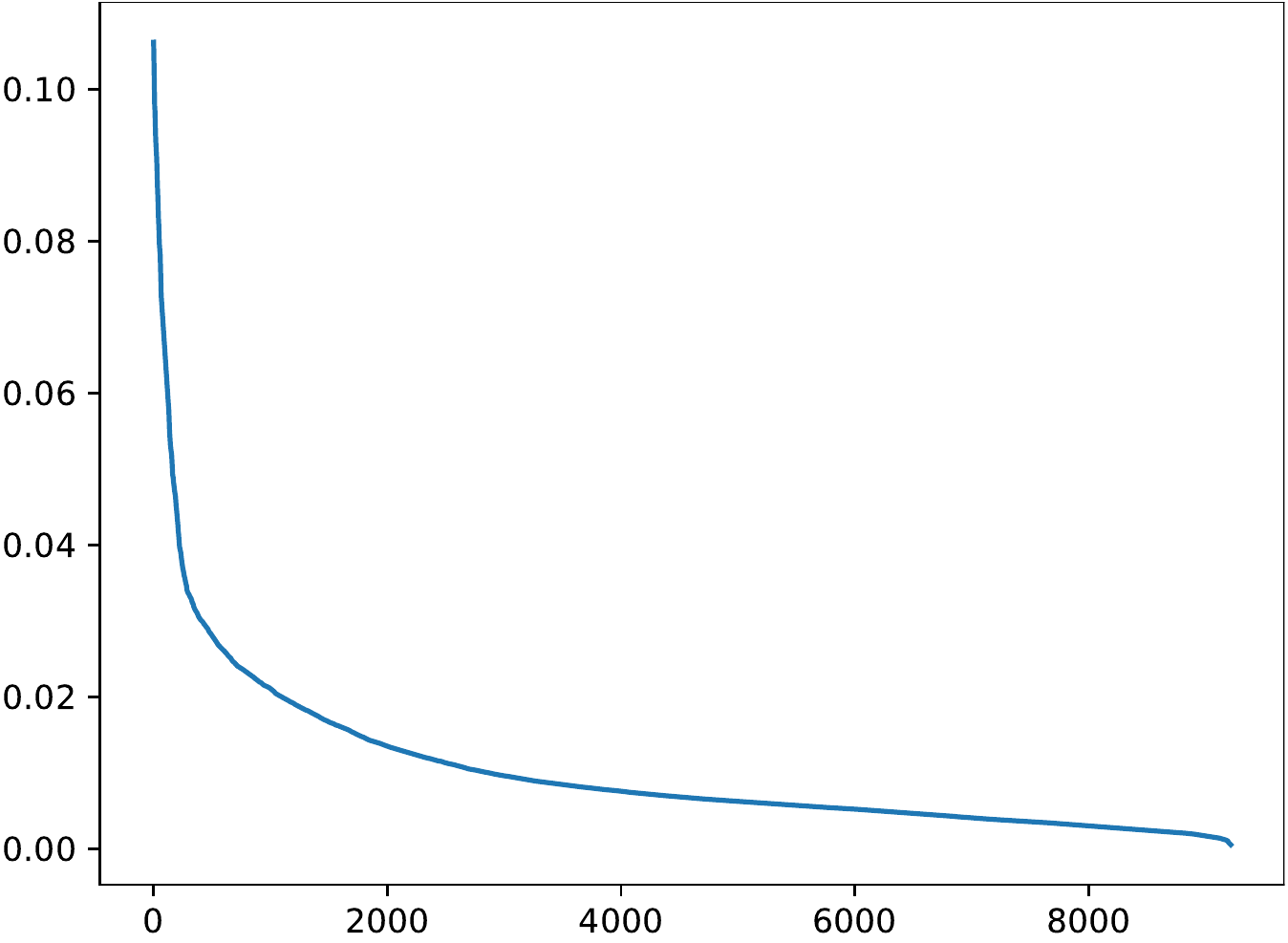}\vspace{1pt} 
		\end{minipage}
		\label{label_for_cross_ref_1}
	}
	\subfigure[Softmax scores at Head7@Encoder layer]{
		\begin{minipage}[t]{0.15\linewidth} 
			\includegraphics[width=\linewidth]{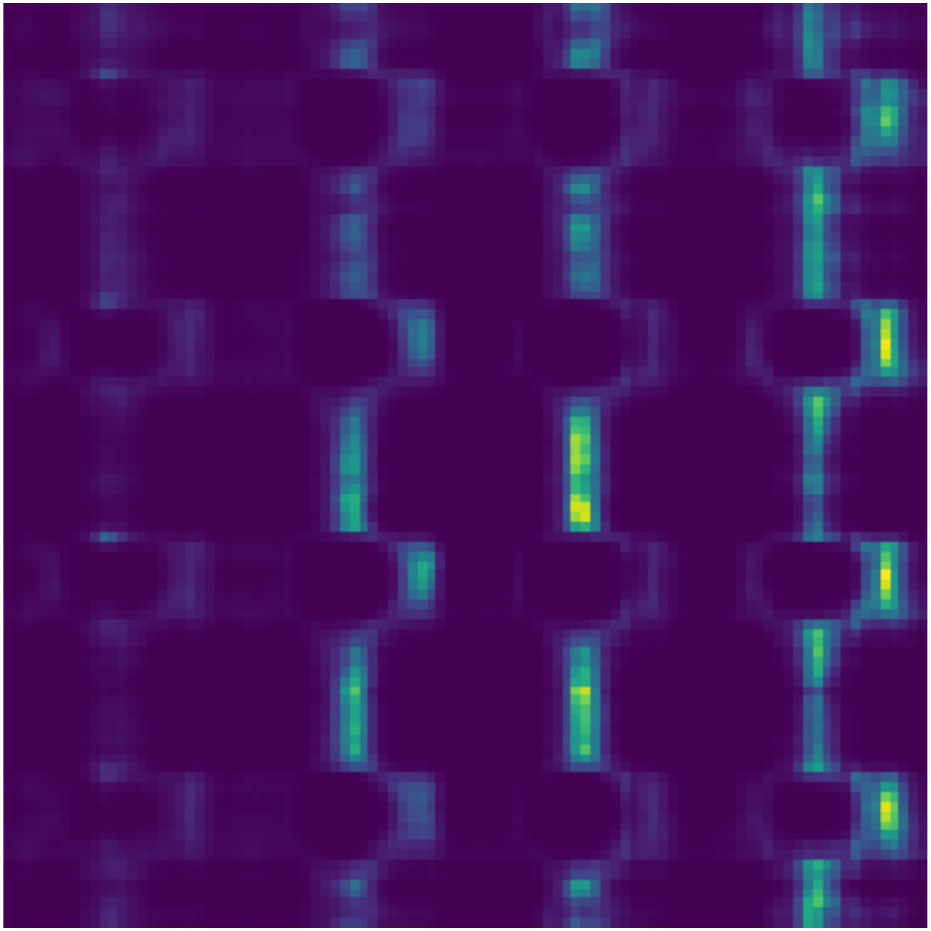}\vspace{1pt}
			\includegraphics[width=\linewidth]{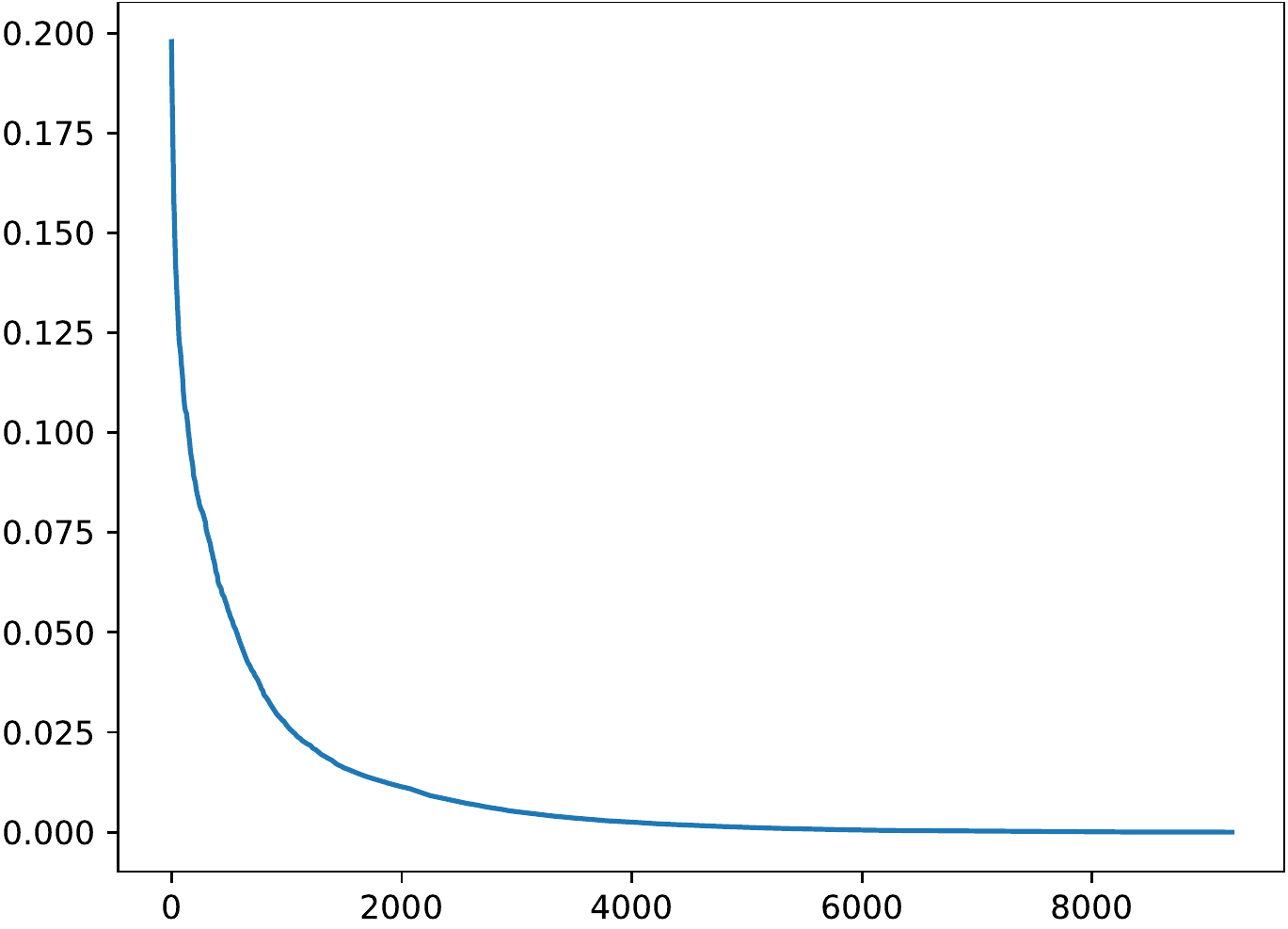}\vspace{1pt} 
		\end{minipage}
		\begin{minipage}[t]{0.15\linewidth}
			\includegraphics[width=\linewidth]{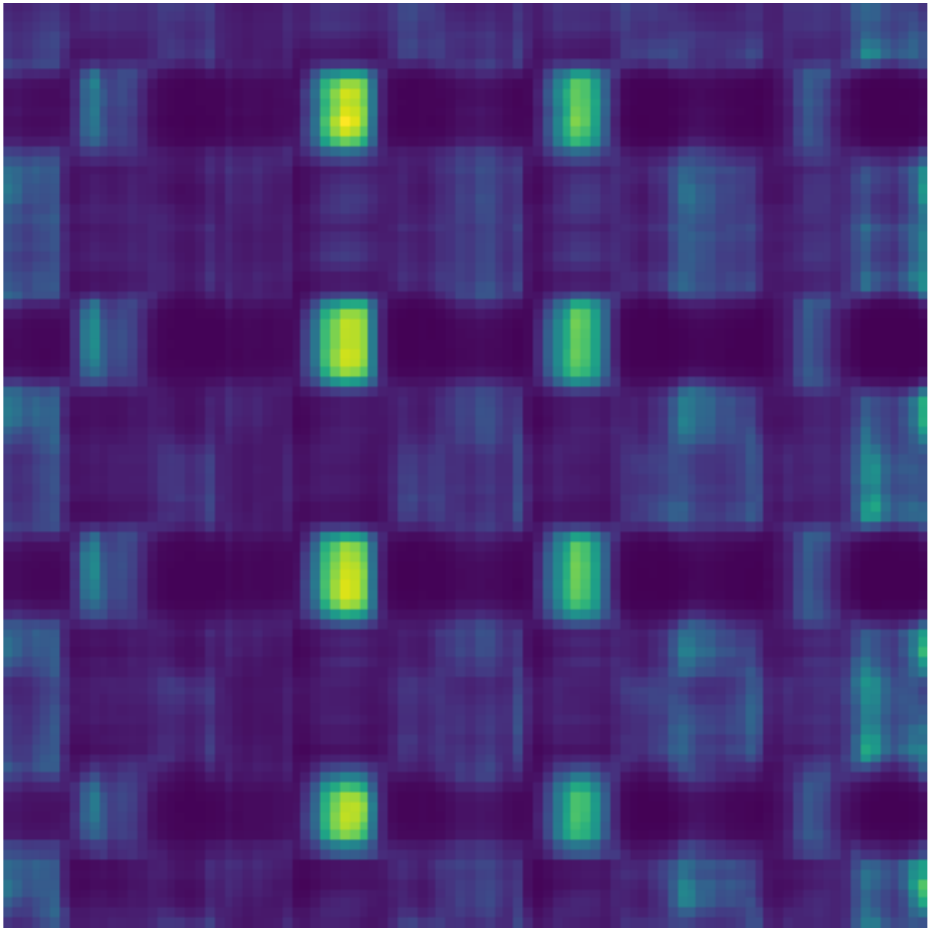}\vspace{1pt} 
			\includegraphics[width=\linewidth]{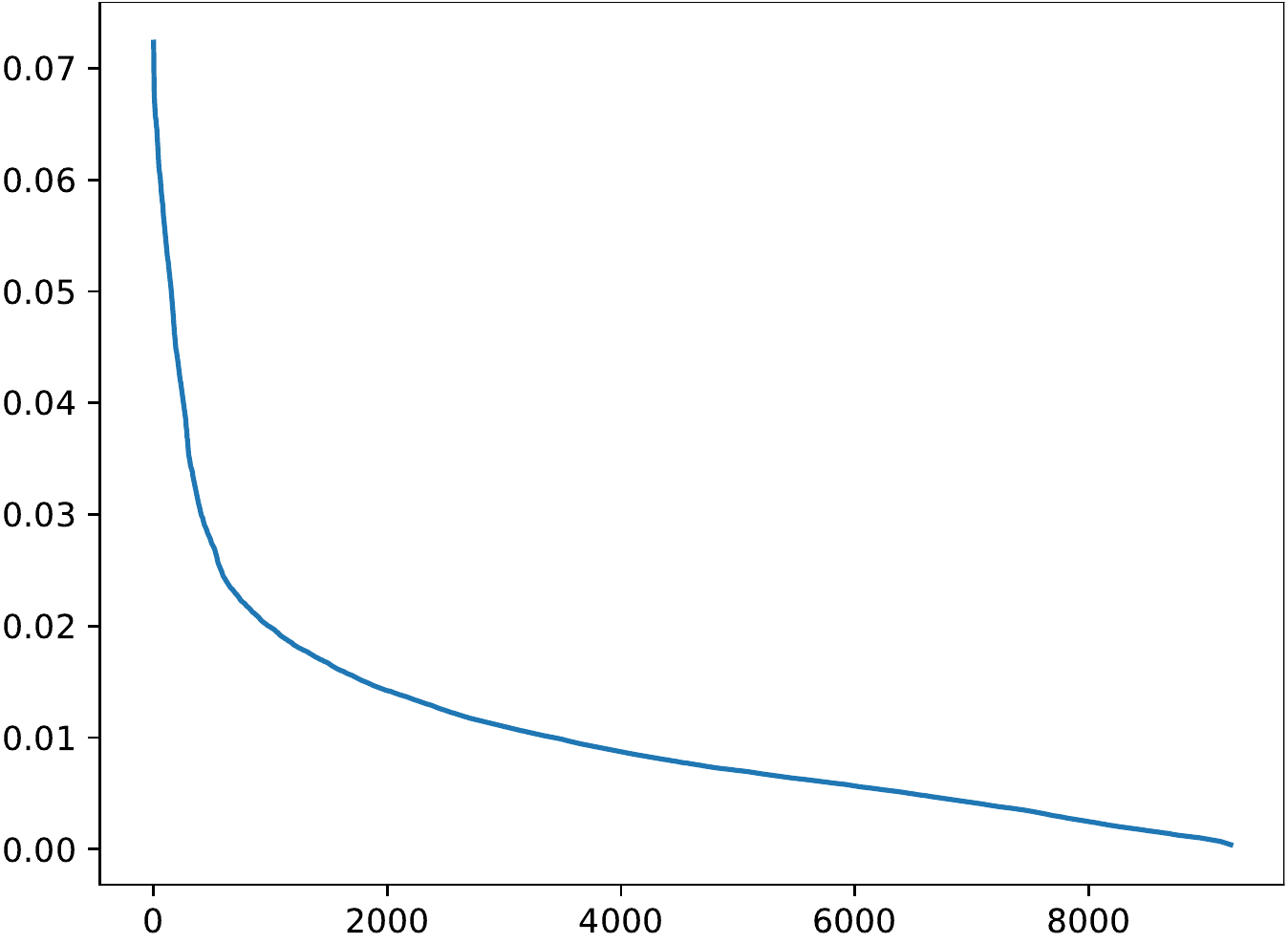}\vspace{1pt}
		\end{minipage}
		\begin{minipage}[t]{0.15\linewidth}
			\includegraphics[width=\linewidth]{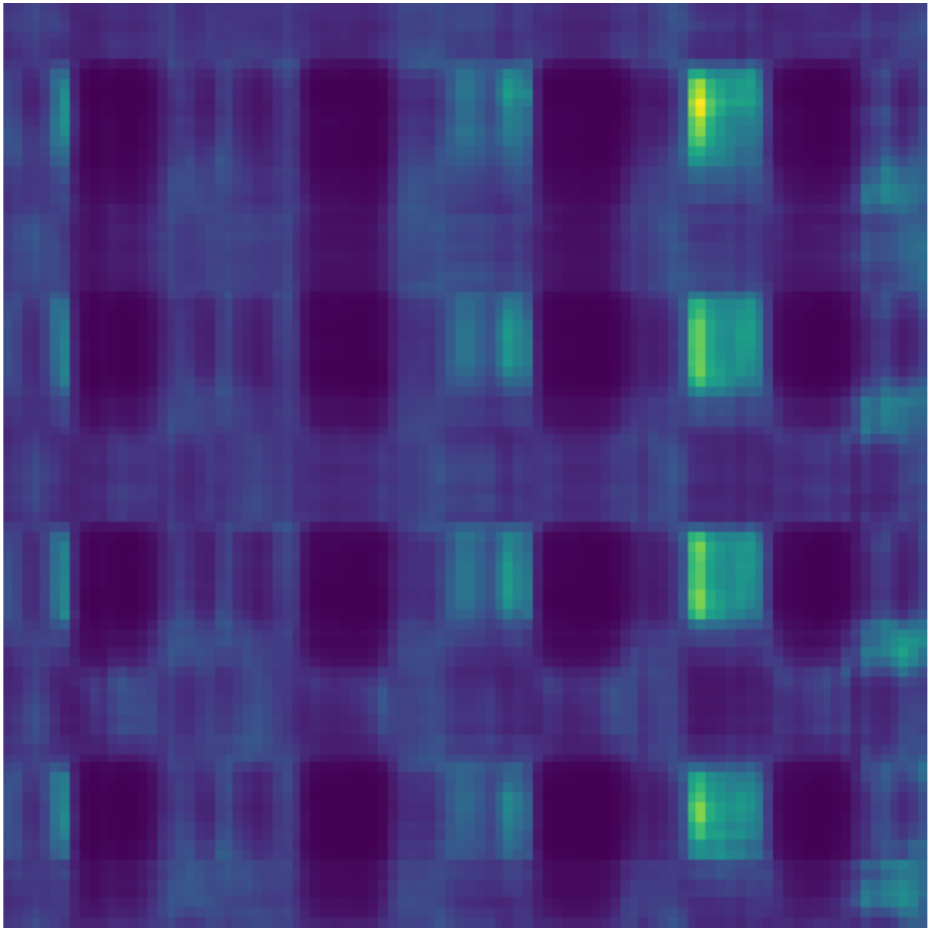}\vspace{1pt} 
			\includegraphics[width=\linewidth]{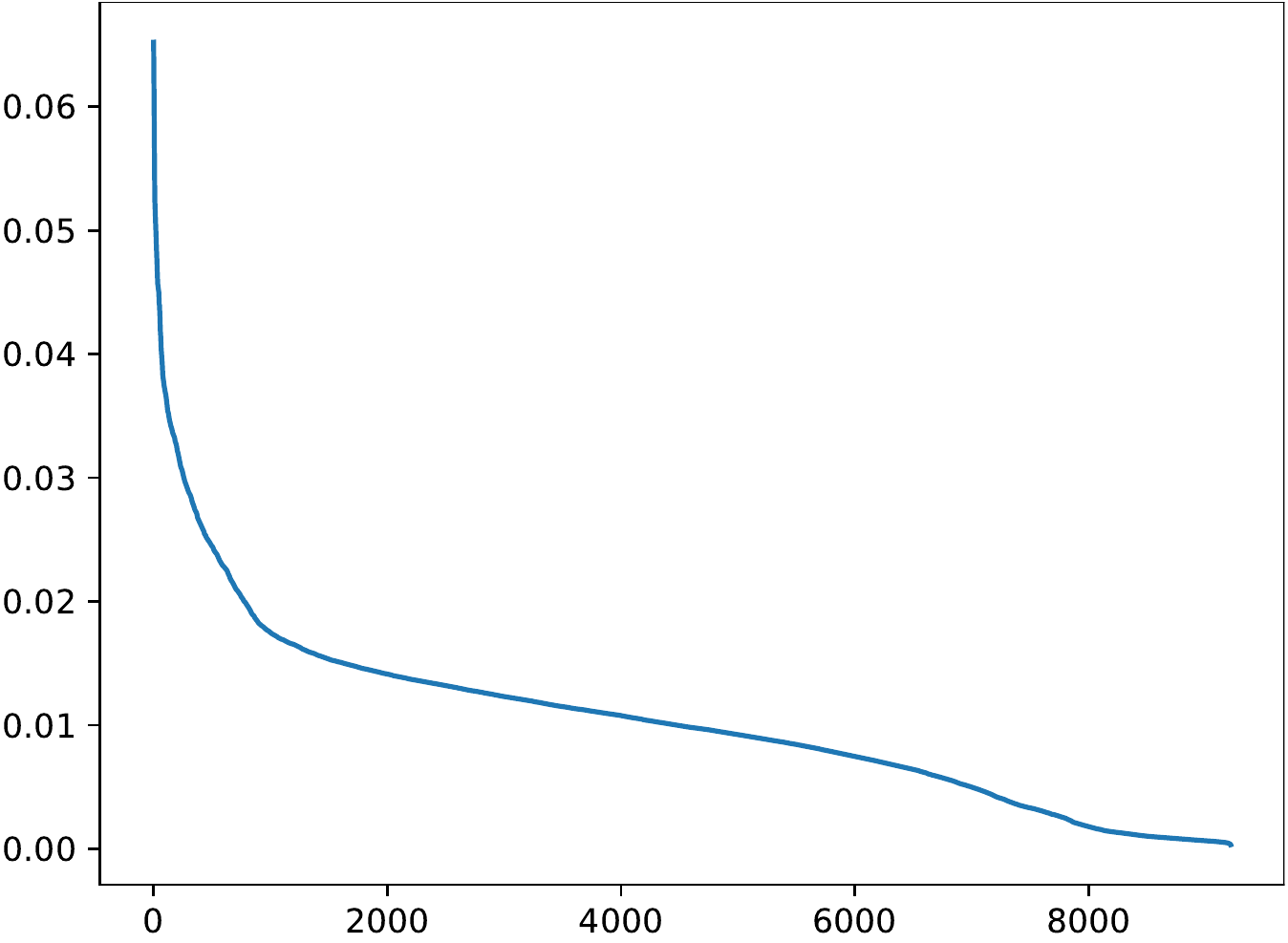}\vspace{1pt} 
		\end{minipage}
		\label{label_for_cross_ref_2}
	}
	\caption{The Softmax scores in the self-attention from canonical Transformer trained on ETTh$_1$ dataset}
	\label{fig1}
\end{figure*}

\section{Related Work}
\subsection{Time-Series Forecasting}
	The classical convolutional neural network (CNN) \cite{krizhevsky2012imagenet} model can extract the local information unrelated to the spatial position in the data \cite{lim2021time}. In order to allow CNN to be used in the time-series, scholars designed multi-layer causal convolutions to ensure that only past information can be used for prediction \cite{borovykh2017conditional, bai2018empirical}. For the processing of long-term dependencies, the Temporal Convolutional Network (TCN) introduces the dilated convolutions, which changes the interval of original look-back window from 1 to $d_l$, where $d_l$ is a layer-specific division rate. In traditional modeling, recurrent neural networ (RNN) is also widely used in the field of time-series prediction owing to its architecture naturally supports inputs and outputs with sequential relationships \cite{salinas2020deepar, rangapuram2018deep, lim2020recurrent, tang2022features}. The main idea is to use the memory state of RNN neurons to store all past effective information. However, RNN variants may be limited in learning the long-term dependency in the data. Since all the information in the past will decay with time and the difficult for RNN to learn the long-term memory \cite{hochreiter2001gradient}. Long Short-Term Memory networks (LSTM) \cite{hochreiter1997long} introduces some different operation gates to solve this problem, but it does not solve the long-term dependency well. 
	To further these effort, attention mechanism is proposed to help the neural network to learn long-term memory information  \cite{bahdanau2014neural}. In short, the attention mechanism of time-series is to calculate the dynamic weight, find the weighted sum of past hidden states, and predict the output value with the summed state. In this way, the vector used for prediction can contain information that predicts a more informative time point for the current time point \cite{fan2019multi, jiang2022accurate, lim2021temporal}. 
	
\subsection{Sparse Attention}
	
	In the standard self-attention mechanism, each token needs to pay attention to all other tokens \cite{vaswani2017attention}. However, for the trained transformer, the learned attention matrix \textbf{A} is often very sparse across most data points \cite{child2019generating}. Therefore, the computational complexity can be reduced by limiting the number of queries that want to participate in the query-key pairs through the incorporating structural bias. The existing methods can be divided into two categories: position-based and content-based sparse attention \cite{lin2021survey}. In position-based sparse attention, the attention matrix is limited to some predefined patterns \cite{ye2019bp, parmar2018image}. Although these spark patterns change in different ways, some of them can be decomposed into some atomic sparse patterns, e.g., global attention, band attention, dilated attention, random attention, block local attention \cite{guo2019star, beltagy2020longformer}. Many spark patterns include one or more of the above atomic sparse patterns \cite{zaheer2020big}. Another work is to create a sparse graph based on the input content. A simple method is to select keywords that may have a large similarity score with a given query. In order to construct the sparse graph effectively, the maximum inner product search problem can be repeated, i.e, the key with the maximum dot product can be found by a query without calculating all dot-product terms \cite{zhou2021informer, li2019enhancing}. For example, Routing transformer \cite{roy2021efficient} uses K-means clustering to cluster queries and keys on the same group of centroid vectors. Each query only focuses on the keys belonging to the same cluster. Reformer \cite{kitaev2019reformer} uses location sensitive hashing (LSH) to select key-value pairs for each query. The proposed LSH  allows each token to attend only to the tokens in the same hash bucket. In Informer \cite{zhou2021informer}, based on query and key similarity sampling dot-product pairs, ProbSparse self-attention is proposed to reduce the time complexity of Transformer to $\mathcal{O}(L\log L)$ and allows it to accept longer input.

\section{Methodology}


The problem of long time-series forecasting is to input the past sequence $\mathcal{X}=\left \{x_{1}, \cdots ,x_{L_x} |x_{i}\in \mathbb{R}^{d_x} \right \}$, and the output is to predict corresponding future sequence $\mathcal{Y}=\left \{y_{L_x + 1}, \cdots ,y_{L_x+L_y } |y_{i}\in \mathbb{R}^{d_y} \right \}$, where $ L_x$ and $ L_y$ are the lengths of input and output sequences respectively, and $d_x$ and $d_y$ are the feature dimensions of input $\mathcal{X}$ and output $\mathcal{Y}$ respectively. The LTFP encourages a longer input's length $L_x$ and a longer output's length $L_y$ than previous works.

Our proposed Infomaxformer holds the encoder-decoder architecture and combines it with the decomposition structure to solve LTFP. Please refer to Figure \ref{imgforemr} for an overview and the following sections for details.
\subsection{Vanilla Self-attention Mechanism}
The scaled dot-product attention mechanism in original Transformer \cite{vaswani2017attention} performs as:
\begin{equation} Attention(\textbf{Q},\textbf{K},\textbf{V}) = Softmax(\frac{\textbf{Q}\textbf{K}^{T}}{\sqrt{d}})\textbf{V}
\end{equation}
i.e., $Attention$ is defined as an operation of ternary matrix, where $\textbf{Q} (queries)\in\mathbb{R}^{L_Q \times d}$, $\textbf{K} (keys)\in \mathbb{R} ^{L_K \times d}$, $\textbf{V} (values)\in \mathbb{R} ^{L_V \times d}$, and $d$ is the feature dimension. To further discuss the self-attention mechanism, the $Softmax$ function is expanded, and use $q_i$, $k_i$ and $v_i$ to represent the $i$-th row in \textbf{Q}, K and \textbf{V} respectively. For the time-series with input length $L$, $i$ represents the $i$-th data. Therefore, the original self-attention mechanism for the $i$-th data can be expressed as:
\begin{equation} \mathcal{A}(i)
	= \sum_{j}^{L}\frac{e^{\frac{q_{i}k_{j}^{T}}{\sqrt{d}}}}{\sum_{l}^{L}e^{\frac{q_{i}k_{l}^{T}}{\sqrt{d}}}}v_{j}
	=\sum_{j}^{L} \frac{k(q_{i},k_{j})}{\sum_{l}^{L}k(q_{i},k_{l})}v_{j}
\end{equation}
which $ k(q_{i},k_{j}) = exp(q_{i}k_{j}^{T}/\sqrt{d})$ \cite{tsai2019transformer}. 

Let $ p(q_i,k_j)=k(q_{i},k_{j})/\sum_{l}^{L}k(q_{i},k_{l})$, $Attention$ can be abbreviated as:
\begin{equation} \mathcal{A}(i)	= \sum_{j}^{L} p\left ( q_{i} , k_{j}\right ) v_{j} \label{eq}
\end{equation}
where $p\left ( q_{i} , k_{j}\right )$ is the probability of $v_i$, then $\mathcal{A}(i)$ is the expectation of matrix \textbf{V}. For the probability $p(q_i,k_j)$, it requires the quadratic times dot-product computation and $\mathcal{O}(L_QL_K)$ memory usage, which is the main reason why the traditional self-attention mechanism cannot handle long time-series (it is easy to lead to out-of-memory), and also the main disadvantage that limits its prediction ability.

Many previous studies have shown that the probability distribution of self-attention mechanism has potential sparsity \cite{child2019generating, lin2021survey}, and a selection strategy is designed for all $p(q_i,k_j)$ without significantly affecting the performance of the model \cite{ye2019bp, parmar2018image, guo2019star, beltagy2020longformer}.
To motivate our approach, we first revisit the learned attention patterns of the vanilla self-attention and make a qualitative evaluation. 
According to Figure \ref{fig1}, in the first layer of encoder, the scores follows an obvious long tail distribution, and the Softmax scores has obvious blocking phenomenon, especially in the second and third layers.
So a few dot-product pairs contribute to the major attention, and others generate negligible attention. 
Then, how to ``select'' them?

\subsection{Reformulation via the Lens of Information Entropy}

We now provide the intuition to reformulate Equation (\ref{eq}) via the lens of information entropy \cite{shannon1948mathematical}. Information entropy is a basic concept of information theory, which describes the uncertainty of possible events of information sources. Its formula is as follows:
\begin{equation}H\left ( x_i \right )= -\sum_{i= 1}^{L}p(x_i)ln p(x_i) \label{hx}
\end{equation}
where $p(x_i)$ represents the probability that the random event $X$ is $x_i$. Any information has redundancy, which is related to the occurrence probability (uncertainty) of each symbol in the information. 
The probability and the amout of information generated are positively correlated.
Information is used to eliminate random uncertainty, and information entropy is a measure of the amount of information needed to eliminate uncertainty, i.e., the amount of information that an unknown event may contain.

\textbf{Maximum Entropy Principle} \textit{When only some knowledge about the unknown distribution is mastered, the probability distribution with the largest entropy value should be selected \cite{jaynes1957information}.}

It is difficult to determine the probability distribution of random variables. 
Generally, only the average values or the values under certain limited conditions can be measured. 
There can be many (even infinite) distributions that meet the measured values. 
The maximum entropy principle is a criterion for selecting the statistical characteristics of random variables that best meet the objective conditions, also known as the Maximum Information Principle. 
Based on this principle, it is effective to select a distribution with maximum entropy as the distribution of the random variable.

\textbf{Maximum Entropy Self-attention} From Equation (\ref{eq}), the $i$-th query's attention on all the keys are defined as a probability distributions $\textbf{p}_i$ and the output is its composition with values $\textbf{V}$. According to the maximum entropy principle, the dominant dot-product pairs encourage the corresponding entropy of $\textbf{p}_i$ to be maximum. However, the traversing of all the $\textbf{p}_i$ still needs to calculate each dot-product pair, i.e., the time complexity is $\mathcal{O}(L^{2})$. Motivated by this, we propose a very simple but effective approximation method to obtain the query information entropy measurement.

\begin{proposition}
	\label{pro1}
	For all probability distributions $\textbf{p}_i$ and $\textbf{p}_j$, if $\sigma_{p_i} <\sigma_{p_j}$, it can be considered that $H(i)>H(j)$.
\end{proposition}

If the $i$-th query's $ \textbf{p}_i$ gains a smaller variance, its information entropy is larger and has a higher possibility to contain the dominate dot-product pairs. Variance is a measure of the degree of dispersion of a group of data. The variance of data subject to the same distribution is the same, so we only need to randomly sample constant $U$ from \textbf{K} to calculate the variance of the $i$-th query's probability distribution $ \textbf{p}_i$, which only need to calculate $\mathcal{O}(L_{Q})$ dot-product for each query-key lookup and the layer memory usage maintains $\mathcal{O}(L_{Q})$. 
Then, select sparse Top-$u$ from \textbf{Q} as $\bar{\textbf{Q}}$ to calculate the standard dot-product pair, so the time complexity and memory usage maintains $\mathcal{O}(uL_K)$. 
However, the rest of queries can't be left without any calculation.

\begin{theorem}
	\label{th1}
	In the case of discrete sources, for discrete sources with \textbf{L} symbols, the information entropy can reach the maximum value only when they appear with equal probability, that is, the average uncertainty of sources with equal probability distribution is the maximum
\end{theorem}

Based on the proposed measurement and Theorem \ref{th1}, we have the maximum entropy self-attention, i.e., MEA (the pseudo-code is in Appendix): 

\begin{equation}
	\mathcal{A}(i) = \begin{cases}
		\sum_{j}^{L} p\left ( q_{i} , k_{j}\right ) v_{j}& \text{ , if top-u} \\ \\
		\sum_{j}^{L} v_{j} /L & \text{, otherwise}
	\end{cases}
	\label{eq:method}
\end{equation}


\subsection{Embedding Method}

\begin{figure}[h]
	\centering
	\includegraphics{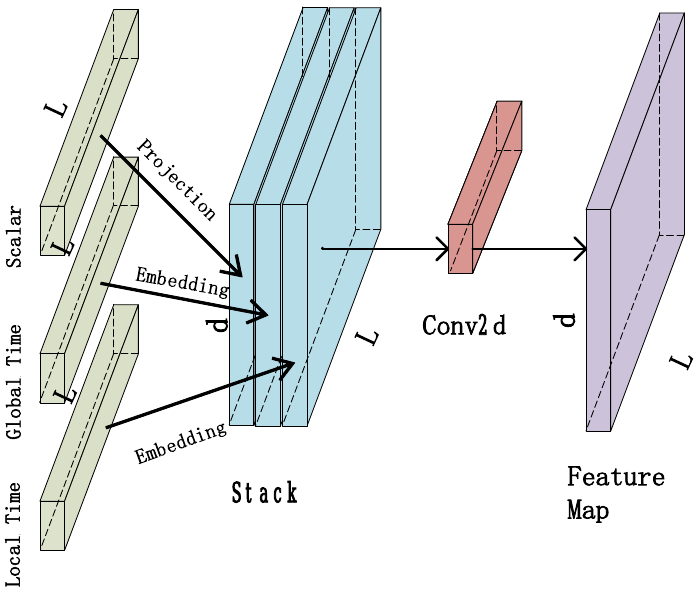}
	\caption{The Embedding Method}
	\label{emb}
\end{figure}

	As shown in Figure \ref{emb}, the input embedding consists of three parts, a scalar, a local position and a global time stamp. We use scalar projection SP, local position embedding PE \cite{vaswani2017attention} and time embedding TE \cite{zhou2021informer} to deal with the above parts respectively:
	\begin{equation}SP = Conv1d(x^t_i) \label{sp} \end{equation} 
	\begin{equation}
		\begin{split}
			PE_{\left ( i, 2j \right )} = sin\left ( i / 10000^{2j/d_{model}}  \right ) \\
			PE_{\left ( i, 2j+1 \right )}= cos\left ( i / 10000^{2j/d_{model}}  \right )
		\end{split}\label{pe}
	\end{equation}
	\begin{equation}TE = E(month) + E(day) + E(hour)+ E(minute) \label{se}
	\end{equation}
	For the Equation (\ref{sp}), we project the scalar context $x^t_i $ into $ d_{model}$-dim vector with 1-D convolutional filters. The kernel width is 3, stride is 1, the input channel is $d_x$ and the output channel is $d_{model}$. For the Equation (\ref{pe}), $ i\in \left \{ 1,\dots,L_x  \right \} $, $ j\in \left \{ 1,\dots,\left \lfloor d_{model}/2 \right \rfloor  \right \} $, $d_{model}$ is the feature dimension after embedding. For the Equation (\ref{se}), $E$ is a learnable stamp embeddings with limited vocab size (up to 60, namely taking minutes as the finest granularity).
	
	For the three different features finally obtained, instead of the method of addition \cite{zhou2021informer, wu2021autoformer}, we stacked them together and reduced their dimension through a two-dimensional convolution, that is, the input channel of convolution is 3 and the output channel is 1:
	\begin{equation}\mathcal{X} =Conv2d(Stack(SP, PE, TE)) 
	\end{equation}
	where the kernel width and stride is $(1,1)$.

\subsection{Keys/Values Distilling}

In previous works, a feed-forward network with a single hidden layer is proposed to linearly project the queries, keys and values \cite{vaswani2017attention, zhou2021informer}. As the natural consequence of the original sequence linear project, the queries, keys and values have a lot of redundant features. We use the distilling operation to privilege the superior keys and values with dominating features and make a focused feature map in the self-attention mechanism. It trims the time dimension of the input sharply, does not arbitrarily delete the feature of the input sequence, but recombines them into a new heads weights matrix. 
\begin{figure}[ht]
	\centering
	\includegraphics[scale=0.9]{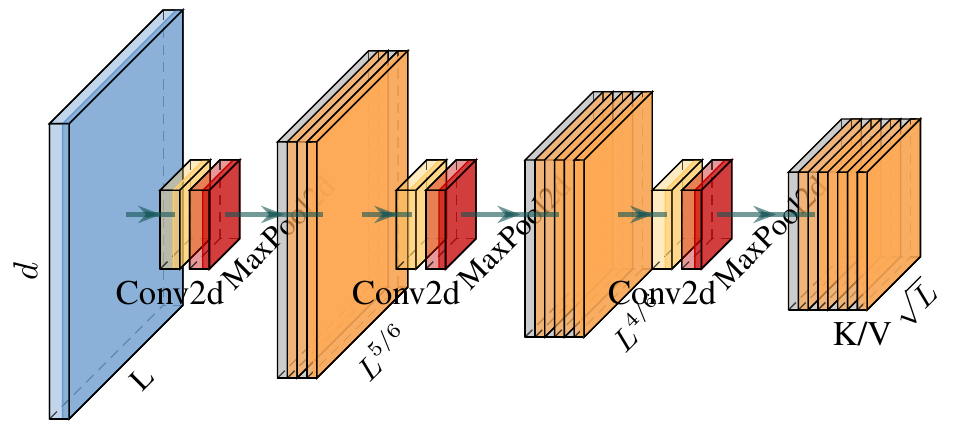}
	\caption{The Keys/Values Distilling}
	\label{imgkv}
\end{figure}

As shown in Figure \ref{imgkv}, we distilling keys/values using a three-step convolution operation:
\begin{equation}
	\begin{split}
		\mathcal{K}_1 &=Maxpool2d(Relu(Conv2d( \mathcal{X})))  \\ 
		\mathcal{K}_2 &=Maxpool2d(Relu(Conv2d( \mathcal{K}_1))) \\ 
		\mathcal{K}_3 &=Maxpool2d(Relu(Conv2d( \mathcal{K}_2)))
	\end{split}
\end{equation}
where $\mathcal{X} \in \mathbb{R}^{L_x\times d_{model}}$. $Conv2d()$ performs an 2-D convolutional filters (kernel size=(2, 2)) with the $Relu$ activation function, the number of input channels is the number of heads, and the number of output channels is $h$ times the number of input channels, so after three-step convolution, the number of output channels, i.e., the number of heads, is $h^3$ (this can be modified as needed). $Maxpool2d()$ performs an 2-D max pooling (kernel size and stride is ($l$, h)). 
Therefore, after three-step convolution, $\mathcal{K}_3 \in \mathbb{R}^{ \frac{L_X}{l^3} \times \frac{d_{model}}{h^3}}$, i.e. $\mathcal{K} \in \mathbb{R}^{L_K \times \frac{d_{model}}{h^3}}$, $\mathcal{V} \in \mathbb{R}^{L_V \times \frac{d_{model}}{h^3}}$, $L_K$ and $L_V$ is ${L_X}/{l^3}$.

\textbf{Complexity Analysis:}
Now we know that $L_K$ is ${L_X}/{l^3}$, so the time complexity and space complexity of our MEA is $\mathcal{O}(u{L_X}/{l^3})$. We set $u=c \sqrt{L_Q}$, $l=L^{1/6}_X$, $c$ is a constant sampling factor, $u$ varies linearly with $L_Q$, so:
\begin{equation}u{L_X}/{l^3}= c \sqrt{L_Q} {L_X}/{(L^{1/6}_X)^3} =  c \sqrt{L_Q} \sqrt{L_X}
\end{equation}
where $L_Q=L_X=L$. Consequently, our time complexity and space complexit can reach linear $\mathcal{O}(L)$.

\subsection{Time-Series Decomposition}

In order to make long-term prediction under the input of long time-series, we use the concept of decomposition to learn complex time patterns, which can separate the time-series into trend and seasonal \cite{hyndman2018forecasting, cleveland1990stl, wu2021autoformer}. These two parts respectively represent two features including long-term development trend and seasonality of the time-series, which are different in different time-series. To overcome such a problem, we introduce a time-series decomposition block (TSD), which can propose the development trend and seasonality of the time-series from the input. Specifically, we use the moving average to smooth out periodic fluctuations, extract long-term trends, and highlight seasonality. For an input sequence $\mathcal{X} \in  \mathbb{R}^{L \times d}$ of length $L$, this process can be formulated as:
\begin{equation}
	\begin{split}
		\mathcal{X}_t & = AvgPool(\mathcal{X}) \\ 
		\mathcal{X}_s &= \mathcal{X} -\mathcal{X}_t
	\end{split}
\end{equation}
where $\mathcal{X}_s, \mathcal{X}_t  \in  \mathbb{R}^{L \times d} $ represent extracted seasonal and trend, respectively. We use $\mathcal{X}_t, \mathcal{X}_s= TimeSeriesDecomp(\mathcal{X})$ to summarize above equations.

\begin{table*}[ht]
	\centering
	\caption{Multivariate long time-series forecasting results on five cases. A lower MSE or MAE indicates a better prediction, and we use black numbers to indicate the best performance. Due to the limitation of memory, the batch size of some models is changed to 16, which is indicated by underlined numbers. The `-' indicates that there is still not enough memory after the batch size is changed to 16}
	\fontsize{9pt}{9pt}\selectfont	
	\resizebox{\linewidth}{!}{	
		\begin{tabular}{cc|cc|cc|cc|cc|cc|cc|cc}
			\toprule[1.0pt]
			\multicolumn{2}{c|}{Methods}                      & \multicolumn{2}{c|}{Infomaxformer} & \multicolumn{2}{c|}{Autoformer}    & \multicolumn{2}{c|}{Informer} & \multicolumn{2}{c|}{Reformer} & \multicolumn{2}{c|}{LogTrans} & \multicolumn{2}{c|}{Transformer} & \multicolumn{2}{c}{LSTM}       \\ 
			\midrule[0.5pt]
			\multicolumn{2}{c|}{Metric}                       & MSE              & MAE             & MSE   & \multicolumn{1}{c|}{MAE}   & MSE           & MAE           & MSE           & MAE           & MSE           & MAE           & MSE             & MAE            & MSE   & MAE                    \\ 
			\midrule[0.5pt]
			\multicolumn{1}{c|}{\multirow{5}{*}{\rotatebox{90}{ECL}}}   & 24  & \textbf{0.190}            & \textbf{0.305 }          & 0.195 & 0.312 & 0.310         & 0.400         & 0.280         & 0.381         & 0.231         & 0.338         & 0.244           & 0.350          & 0.338 & 0.419                  \\
			\multicolumn{1}{c|}{}                       & 48  & \textbf{0.209}            & \textbf{0.321}           & \underline{0.221} & \underline{0.332 }& 0.357         & 0.425         & 0.273         & 0.370         & 0.287         & 0.373         & 0.260           & 0.359          & 0.334 & 0.412                  \\
			\multicolumn{1}{c|}{}                       & 96  & \textbf{0.218}            & \textbf{0.328}           & \underline{0.230} & \underline{0.340} & 0.367         & 0.434         & 0.291         & 0.381         & 0.292         & 0.377         & 0.278           & 0.373          & 0.330 & 0.409                 \\
			\multicolumn{1}{c|}{}                       & 192 & \textbf{0.239}            &\textbf{ 0.346}           & \underline{0.280} & \underline{0.347} & 0.362         & 0.434         & 0.344         & 0.420         & 0.295         & 0.385         & \underline{0.286}          & \underline{0.377}         & 0.327 & 0.407                  \\
			\multicolumn{1}{c|}{}                       & 384 & \textbf{0.278}            & \textbf{0.377}           & \multicolumn{2}{c|}{-}             & 0.450         & 0.483         & \underline{0.327}         & \underline{0.404}         & 0.323         & 0.397         & \underline{0.290}          & \underline{0.378}         & 0.320 & 0.403                  \\ 
			\midrule[0.5pt]
			\multicolumn{1}{c|}{\multirow{5}{*}{\rotatebox{90}{ETTh$_1$}}} & 24  & \textbf{0.478}            & \textbf{0.489}           & 0.499 & 0.516 & 1.130         & 0.871         & 0.624         & 0.578         & 0.505         & 0.513         & 0.882           & 0.723          & 1.232 & 0.839                  \\
			\multicolumn{1}{c|}{}                       & 48  & \textbf{0.552}            & \textbf{0.528}          & \underline{0.562} & \underline{0.552} & 1.231         & 0.905         & 0.727         & 0.638         & 0.568         & 0.544         & 1.311           & 0.954          & 1.261 & 0.873                 \\
			\multicolumn{1}{c|}{}                       & 96  & \textbf{0.564}           & \textbf{0.543}          & \underline{0.611} & \underline{0.579} & 1.345         & 0.953         & 0.930         & 0.743         & 0.714         & 0.629         & 1.957           & 1.199          & 1.268 & 0.873                 \\
			\multicolumn{1}{c|}{}                       & 192 & \textbf{0.556}            & \textbf{0.544}           & \underline{0.724} & \underline{0.625} & 1.643         & 1.061          & 1.124         & 0.821         & 0.865         & 0.717         & \underline{1.758}          & \underline{1.136}          & 1.266 & 0.871                \\
			\multicolumn{1}{c|}{}                       & 384 & \textbf{0.597}            & \textbf{0.570}           & \multicolumn{2}{c|}{-}              & 1.499         & 1.004         & \underline{1.270}         & \underline{0.862}         & 0.952         & 0.749         & \underline{1.405}           & \underline{0.981}          & 1.273 & 0.873                 \\
			\midrule[0.5pt]
			\multicolumn{1}{c|}{\multirow{5}{*}{\rotatebox{90}{ETTh$_2$}}} & 24  & \textbf{0.436}            & \textbf{0.476 }          & 0.437 & 0.493 & 2.048         & 1.173         & 0.975         & 0.787         & 0.621         & 0.617         & 1.016           & 0.814          & 3.291    &1.385 \\
			\multicolumn{1}{c|}{}                       & 48  & \textbf{0.629}            & \textbf{0.544}          &\underline{ 0.658} & \underline{0.643} & 3.047         & 1.471         & 1.652         & 1.027         & 1.168         & 0.985         & 2.199           & 1.242          & 3.378   & 1.408 \\
			\multicolumn{1}{c|}{}                       & 96  & \textbf{0.593}            & \textbf{0.533}           & \underline{0.686} & \underline{0.646} & 6.882         & 2.258         & 3.301         & 1.427         & 2.279         & 1.265         & 5.862           & 2.052          & 3.488    & 1.432 \\
			\multicolumn{1}{c|}{}                       & 192 & \textbf{0.703}            & \textbf{0.607}           & \underline{0.792} & \underline{0.671} & 5.070         & 1.885         & 3.774         & 1.617         & 4.207         & 1.776         & \underline{4.045}          &\underline{1.675}         & 3.489    & 1.434 \\
			\multicolumn{1}{c|}{}                       & 384 & \textbf{0.575}            & \textbf{0.557}           & \multicolumn{2}{c|}{-}             & 4.080         & 1.669         & \underline{3.363}         & \underline{1.465 }        & 3.032         & 1.526         & \underline{3.549}           & \underline{1.516}          & 3.486   & 1.430\\ 
			\midrule[0.5pt]
			
			\multicolumn{1}{c|}{\multirow{5}{*}{\rotatebox{90}{ETTm$_1$}}} & 24  &\textbf{0.330}           & \textbf{0.397}           & 0.414 & 0.438 & 0.354         & 0.401         & 0.430         & 0.453         & 0.882         & 0.666         & 0.355           & 0.403          & 1.121 & 0.791                 \\
			\multicolumn{1}{c|}{}                       & 48  & \textbf{0.418}            & \textbf{0.454}           &\underline{0.537} & \underline{0.505} & 0.533         & 0.521         & 0.578         & 0.544         & 0.951         & 0.707         & 0.514           & 0.526          & 1.130 & 0.799                 \\
			\multicolumn{1}{c|}{}                       & 96  & \textbf{0.494}            & \textbf{0.502}           & \underline{0.545} & \underline{0.517} & 0.592         & 0.571         & 0.710         & 0.612         & 0.558         & 0.540         & 0.740           & 0.657          & 1.141 & 0.805                 \\
			\multicolumn{1}{c|}{}                       & 192 & \textbf{0.558}            & \textbf{0.531}         & \underline{0.605} & \underline{0.534} & 0.768         & 0.682         & 0.896         & 0.702         & 0.591         & 0.565         & \underline{0.700}          & \underline{0.641}       & 1.141 & 0.806                      \\
			\multicolumn{1}{c|}{}                       & 384 & \textbf{0.611}          &\textbf{0.561}           & \multicolumn{2}{c|}{-}             & 0.938         & 0.765         & \underline{1.072}       &\underline{ 0.781}         & 0.767         & 0.650         & \underline{0.838}           & \underline{0.719}          & 1.153 &0.810                  \\ 
			\midrule[0.5pt]
			
			\multicolumn{1}{c|}{\multirow{5}{*}{\rotatebox{90}{Weather}}}   & 24  &\textbf{0.307}            & \textbf{0.357}           & 0.455 & 0.489 & 0.348         & 0.401         & 0.370           & 0.426            & 0.385         & 0.427         & 0.326           & 0.378          & 0.492 & 0.500                 \\
			\multicolumn{1}{c|}{}                       & 48  & \textbf{0.381}           & \textbf{0.422}           & \underline{0.544} & \underline{0.542} & 0.488         & 0.505         & 0.443            & 0.475            & 0.498         & 0.505         & 0.447           & 0.664          & 0.497 & 0.504                \\
			\multicolumn{1}{c|}{}                       & 96  &\textbf{0.456}           &\textbf{0.481}          &\underline{0.554} & \underline{0.546} & 0.603         & 0.574         & 0.511            & 0.520            & 0.562         & 0.550         & 0.548           & 0.532          & 0.500 & 0.506                  \\
			\multicolumn{1}{c|}{}                       & 192 & \textbf{0.508}           &\textbf{0.517}           &\underline{0.585} & \underline{0.560} & 0.700         & 0.632         & 0.537            & 0.541            & 0.591         & 0.570         & \underline{0.620}           &\underline{0.575}         & 0.503 & 0.517                  \\
			\multicolumn{1}{c|}{}                       & 384 & \textbf{0.511}            & \textbf{0.514}          & \multicolumn{2}{c|}{-}             & 0.681         & 0.621         & \underline{0.536}  &\underline{0.535}        & 0.622         & 0.585         &\underline{0.630}          & \underline{0.579}          & 0.513 & 0.524        \\ 
			\bottomrule[1.0pt] 
		\end{tabular}
	}
	
	\label{tabmlst}
\end{table*}

\subsection{Encoder and Decoder}
$\\$

\textbf{Encoder:}
As shown in Figure \ref{imgforemr}, the encoder focuses on modeling of the seasonal part. Our encoder layers are composed of two sub-blocks. The first is a MEA mechanism, and the second is a simple, position-wise fully connected feed-forward network (MLP). We employ residual connections \cite{he2016deep} around each of the sub-blocks, but unlike previous structures \cite{vaswani2017attention, 2021An}, layer normalization \cite{ba2016layer} was not performed. The input of the encoder is only the seasonal part $\mathcal{X}_s$ of the input sequence $\mathcal{X}$, and the output only contains the seasonal information of the past and will be used as cross information to help the decoder better predict the seasonal information of the future sequence.
\begin{equation}
	\begin{split}
		\mathcal{X}_t, \mathcal{X}_s&= TimeSeriesDecomp(\mathcal{X})\\ 
		\mathcal{X}^n_{s1} &= \mathcal{X}^{n-1}_s + MEA(\mathcal{X}^{n-1}_s)\\
		\mathcal{X}^n_{s}&	=\mathcal{X}^n_{s1} + MLP(\mathcal{X}^n_{s1})
	\end{split}
\end{equation}
where  $n = 1 \dots N $, $\mathcal{X}^0_t =\mathcal{X}_t $, $ \mathcal{X}_{eno} = \mathcal{X}^N_{t}$, $N$ is the number of layers of the encoder, and MLP consists of two 1-D convolution operations.

\textbf{Decoder:}
In addition to the two sub-blocks in each encoder layer, the decoder in classic Transformer also inserts a third sub-block in the two sub-blocks, which performs self-attention on the output of the encoder layers. On this basis, we insert the fourth sub-block in the decoder, i.e., the time-series decomposition block. Similar to the encoder, we employ residual connections
around each of the sub-blocks, but did not perform layer normalization. We input the following vectors to the decoder:
\begin{equation}	\mathcal{X}_{dei} = Concat(	\mathcal{X}_{label},\mathcal{X}_{0})  \in  \mathbb{R}^{(L_{label}+L_y) \times d_{model}} 
\end{equation}
where $ \mathcal{X}_{label}  \in  \mathbb{R}^{L_{label} \times d_{model}}$ is start token, $L_{label}$ is the label length, 
$ \mathcal{X}_{0}  \in  \mathbb{R}^{L_{y} \times d_{model}}$ is a placeholder for the target sequence (set the scalar to 0).
By setting masked dot-products to negative infinity, masked multi-head attention is applied to the MEA calculation (MMEA). This masking ensures that the prediction of position $i$ can only rely on the known outputs of positions less than $i$, which avoids auto-regressive. A fully connected layer acquires the final output, and its outsize is $d_y$. For the time-series, the trend changes are not obvious, but the specific seasonal is different. We use the decoder to predict the seasonal of future data, and use the average of input data to approximate the trend part of future data.
\begin{equation}
	\begin{split}
		\mathcal{X}^n_1& = \mathcal{X}^{n-1} + MEA(\mathcal{X}^{n-1})\\
		\mathcal{X}^n_{t},\mathcal{X}^n_{s}& = TimeSeriesDecomp(\mathcal{X}^n_1)\\
		\mathcal{X}^n_{s1}& = \mathcal{X}^n_{s} +  MMEA(\mathcal{X}^n_{s},\mathcal{X}_{eno})\\
		\mathcal{X}^n &=\mathcal{X}^n_{s1} + MLP(\mathcal{X}^n_{s1}) + \mathcal{X}^n_{t}\\
		\mathcal{Y}& = Mean(\mathcal{X}_t) + \mathcal{X}^M
	\end{split}
\end{equation}
where  $n = 1 \dots M $, $\mathcal{X}^0 =\mathcal{X}_{dei}$, $M$ is the number of layers of the decoder. 

\textbf{Loss Function:} Our loss function is calculated by the mean square error (MSE) between the model output data $y_{o}$ and the real data $y$. and the loss is propagated back from the decoder’s outputs across the entire model.


\begin{table}[h]
	\centering
	\caption{Complexity analysis of different forecasting models. The ${\star}$ denotes applying generative style decoder \cite{zhou2021informer}.}
	\begin{tabular}{l|cc|c}
		\toprule[1.0pt]
		\multicolumn{1}{c|}{\multirow{2}{*}{Methods}} & \multicolumn{2}{c|}{Training}     & Forecasting \\ \cmidrule{2-4} 
		\multicolumn{1}{c|}{}                         & Time                     & Memory & Steps   \\
		\midrule[0.5pt]
		Infomaxformer                                 & \multicolumn{1}{c|}{$\mathcal{O}(L)$} & $\mathcal{O}(L)$   & 1       \\
		\midrule[0.5pt]
		Autoformer                                    & \multicolumn{1}{c|}{$\mathcal{O}(L \log L)$} & $\mathcal{O}(L\log L)$    & 1       \\
		\midrule[0.5pt]
		Informer                                      & \multicolumn{1}{c|}{$\mathcal{O}(L\log L)$} & $\mathcal{O}(L\log L)$    & 1       \\
		\midrule[0.5pt]
		Transformer                                   & \multicolumn{1}{c|}{$\mathcal{O}(L^2)$} & $\mathcal{O}(L^2)$   & $L$       \\
		\midrule[0.5pt]
		LogTrans                                      & \multicolumn{1}{c|}{$\mathcal{O}(L\log L)$} & $\mathcal{O}(L^2)$   & 1$^{\star}$        \\
		\midrule[0.5pt]
		Reformer                                      & \multicolumn{1}{c|}{$\mathcal{O}(L\log L)$} & $\mathcal{O}(L\log L)$    & $L$       \\
		\midrule[0.5pt]
		LSTM                                          & \multicolumn{1}{c|}{$\mathcal{O}(L)$} & $\mathcal{O}(L)$    &$L$     \\
		\bottomrule[1.0pt] 
	\end{tabular}
	
	\label{tabcomplexity}
\end{table}

\begin{table*}[ht]
	\centering
	\caption{Different input lengths for two prediction lengths in ETTh$_1$. The `-' indicates failure for the out-of-memory}
	\resizebox{\linewidth}{!}{
		\begin{tabular}{cc|cccccc|ccccc}
			\toprule[1.0pt]
			\multicolumn{2}{c|}{Predicition length} & \multicolumn{6}{c|}{336}                                                                                         & \multicolumn{5}{c}{480}                                                                            \\
			\midrule[0.5pt]
			\multicolumn{2}{c|}{Encoder's input}    & 336            & 480            & 720            & 960                & 1200               & 1440               & 480            & 720                & 960                & 1200               & 1440               \\
			\midrule[0.5pt]
			\multirow{2}{*}{Informer}       & MSE  & 1.474          & 1.576          & 1.644          & 1.607              & 1.580              & 1.766              & 1.441          & 1.531              & 1.508              & 1.428              & 1.520              \\
			& MAE  & 0.999          & 1.024          & 1.045          & 1.043              & 1.037              & 1.124              & 0.971          & 0.998              & 0.997              & 0.967              & 1.017              \\
			\midrule[0.5pt]
			\multirow{2}{*}{Reformer}       & MSE  & 1.000          & 1.043          & 1.200          & 1.159              & \multirow{2}{*}{-} & \multirow{2}{*}{-} & 1.148          & 1.259              & \multirow{2}{*}{-} & \multirow{2}{*}{-} & \multirow{2}{*}{-} \\
			& MAE  & 0.766          & 0.790          & 0.842          & 0.829              &                    &                    & 0.827          & 0.870              &                    &                    &                    \\
			\midrule[0.5pt]
			\multirow{2}{*}{Transformer}       & MSE  & 1.085          & 1.161          & 1.630          & 1.922              & \multirow{2}{*}{-} & \multirow{2}{*}{-} & 1.083          & 1.410              & \multirow{2}{*}{-} & \multirow{2}{*}{-} & \multirow{2}{*}{-} \\
			& MAE  & 0.830          & 0.858          & 1.065          & 1.159             &                    &                    & 0.836          & 0.969              &                    &                    &                    \\
			\midrule[0.5pt]
			\multirow{2}{*}{LogTrans}       & MSE  & 1.136          & 1.127          & 1.059          & 1.075              & 1.060              & 1.067              & 0.888          & 0.940              & 0.892              & 0.885              & 0.914              \\
			& MAE  & 0.851          & 0.849          & 0.817          & 0.823              & 0.818              & 0.816              & 0.733          & 0.764              & 0.736              & 0.732              & 0.745              \\
			\midrule[0.5pt]
			\multirow{2}{*}{Autoformer}     & MSE  & 0.581          & 0.560          & 0.748          & \multirow{2}{*}{-} & \multirow{2}{*}{-} & \multirow{2}{*}{-} & 0.573         & \multirow{2}{*}{-} & \multirow{2}{*}{-} & \multirow{2}{*}{-} & \multirow{2}{*}{-} \\
			& MAE  & 0.547          & 0.547          & 0.650          &                    &                    &                    & 0.558         &                    &                    &                    &                    \\
			\midrule[0.5pt]
			\multirow{2}{*}{Infomaxformer}  & MSE  & \textbf{0.567} & \textbf{0.556} & \textbf{0.544} & \textbf{0.566}     & \textbf{0.568}     & \textbf{0.624}     & \textbf{0.537} & \textbf{0.583}     & \textbf{0.597}     & \textbf{0.599}     & \textbf{0.709}     \\
			& MAE  & \textbf{0.545} & \textbf{0.547} & \textbf{0.543} & \textbf{0.563}     & \textbf{0.564}     & \textbf{0.602}     & \textbf{0.551} & \textbf{0.581}     & \textbf{0.587}     & \textbf{0.585}     & \textbf{0.653}   \\ 
			\bottomrule[1.0pt] 
		\end{tabular}
	}

	\label{tablength}
\end{table*}

\section{Experiment}

\subsection{Datasets}\label{SCM}
$\\$
\textbf{ETT(Electricity Transformer Temperature):}\footnote{ETT dataset was acquired at \cite{zhou2021informer}.}
The ETT is a key indicator for long-term deployment of the electric power, which collects data from two counties in China from July 2016 to July 2018 for a total of two years.

\textbf{ECL (Electricity Consuming Load):}\footnote{ECL dataset was acquired at \url{https://archive.ics.uci.edu/ml/datasets/ElectricityLoadDiagrams20112014}.}
It collects the electricity consumption (Kwh) of 321 customers. Due to the lack of data \cite{li2019enhancing}, we follow the settings in Informer \cite{zhou2021informer} to convert the dataset to hourly consumption for 2 years.

\textbf{Weather:}\footnote{Weather dataset was acquired at \url{https://www.ncei.noaa.gov/data/local-climatological-data/}.}
This dataset contains the local climatologicale data of nearly 1600 locations in the United States. The data are collected by once an hour from 2010 to 2013.

\subsection{Experimental Details}$\\$
\textbf{Baselines:}
We selected six methods as comparison, including Transformer \cite{vaswani2017attention}, four latest state-of-the-art Transformer-based models: Reformer \cite{kitaev2019reformer}, LogTrans \cite{li2019enhancing}, Informer \cite{zhou2021informer}, Autoformer \cite{wu2021autoformer}, and one RNN-based models: LSTM (Long Short-Term Memory networks) \cite{hochreiter1997long}.

\textbf{Experiment setting:}
Our experiment was implemented in Pytoch \cite{paszke2019pytorch}, and all the experiments are conducted on a single Nvidia RTX 3090 GPU (24GB memory). The input of each dataset is zero-mean normalized. We use two evaluation metrics, including mean square error (MSE): $MSE= \frac{1}{n}  \sum_{i=1}^{n} \sum_{j=1}^{d} \frac{(y-\hat{y} )^2}{d}$ and mean absolute error (MAE): $MAE= \frac{1}{n}  \sum_{i=1}^{n} \sum_{j=1}^{d} \frac{\left | y-\hat{y}  \right |}{d} $,
where $n$ is the length of the sequence and $d$ is the dimension of data at each time point. We use these two evaluation metrics on each prediction window to calculate the average of forecasts and roll the whole set with $stride=1$.

Our implementation details follows common practice of Informer \cite{zhou2021informer} training, and all experiments are repeated five times. We use Adam \cite{kingma2014adam} optimizer for optimization with a learning rate starts from $1e^{-4}$, decaying two times smaller every epoch, and the batch size is 32. The number of encoder layers is 3 and the number of decoder layers is 2. There is no limit to the total number of epochs, with appropriate early stopping, i.e., when the loss of the validation set does not decrease on three epochs, the training will be stopped. More detailed settings can be found in Appendix 4.2.

\subsection{Multivariate Time-series Forecasting}

To compare the performance of different prediction lengths, we fixed the input length $L_x$ to 784 and gradually extended the prediction length $L_y$,  i.e., \{24, 48, 96, 192, 384\}, representing \{6h, 12h, 24h, 48h, 96h\} in ETTm, \{1d, 2d, 4d, 8d, 16d\} in \{ETTh, ECL, Weather\}, and we set the length of the label to double $L_y$.

As shown in Table \ref{tabmlst}, our proposed Infomaxformer model achieves the best performance in all benchmarks and all predicted length settings. Although the performance of Autoformer is closest to our model, it can only set the batch size to 32 when the prediction length is 24. When the prediction length is 384, the batch size to 16 will also lead to out-of-memory.
This shows that our proposed Infomaxformer model can increase the prediction ability, while greatly reducing the use of memory.  
In addition, we also found that with the increase of prediction length, the prediction performance of Infomaxformer is more stable, and there is no sudden drop in performance, which means that Infomaxformer maintains good long-term robustness. Low memory usage, good robustness, high-performance prediction, etc., which is very meaningful for practical applications, and our model has the above advantages.

\begin{figure*}[htbp]
	\centering
	\subfigure[Sampling Factor $c$]{
		\begin{minipage}[t]{0.23\linewidth} 
			\includegraphics[width=\linewidth]{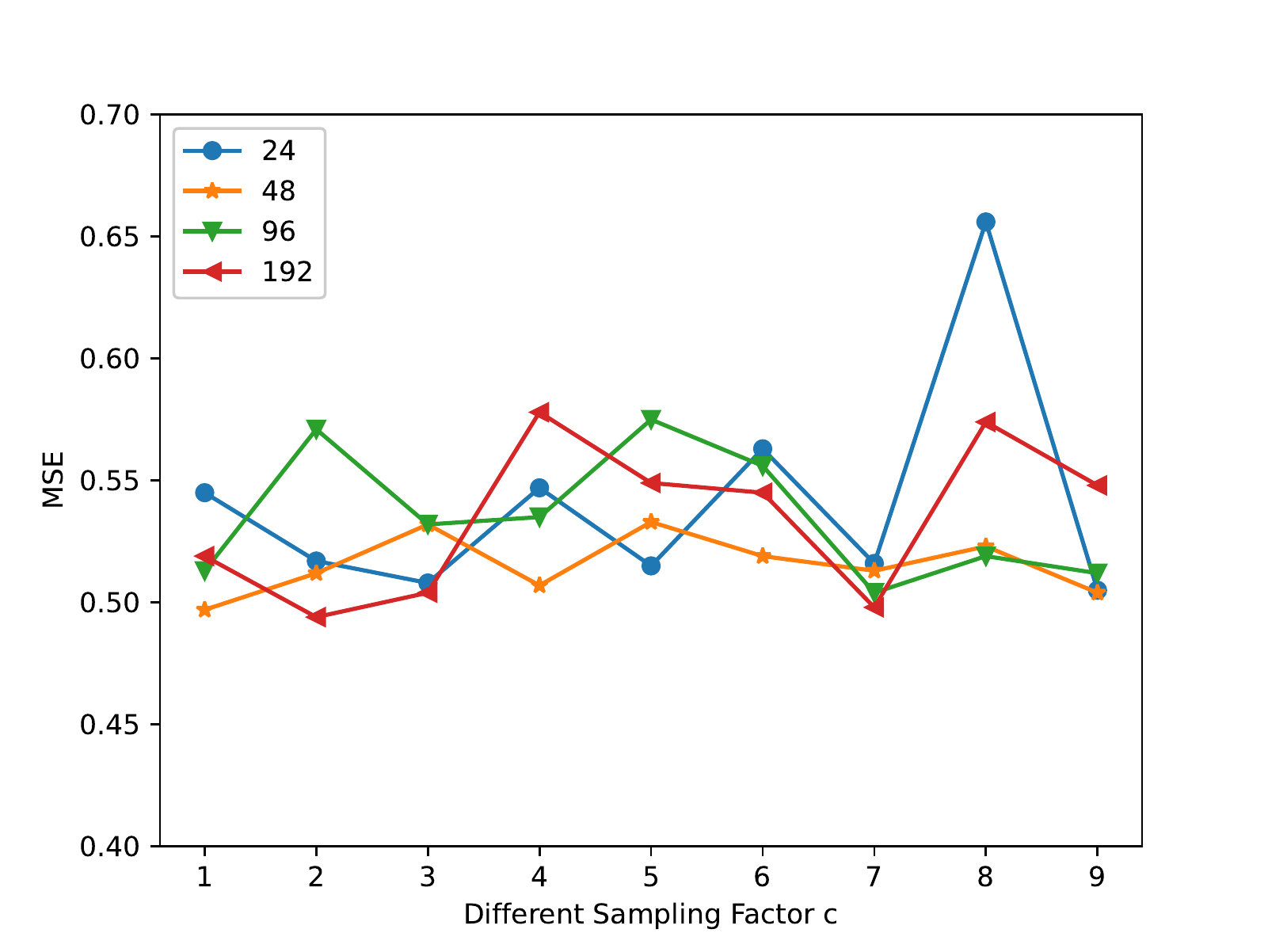}\vspace{1pt}  
		\end{minipage}
		\begin{minipage}[t]{0.23\linewidth} 
			\includegraphics[width=\linewidth]{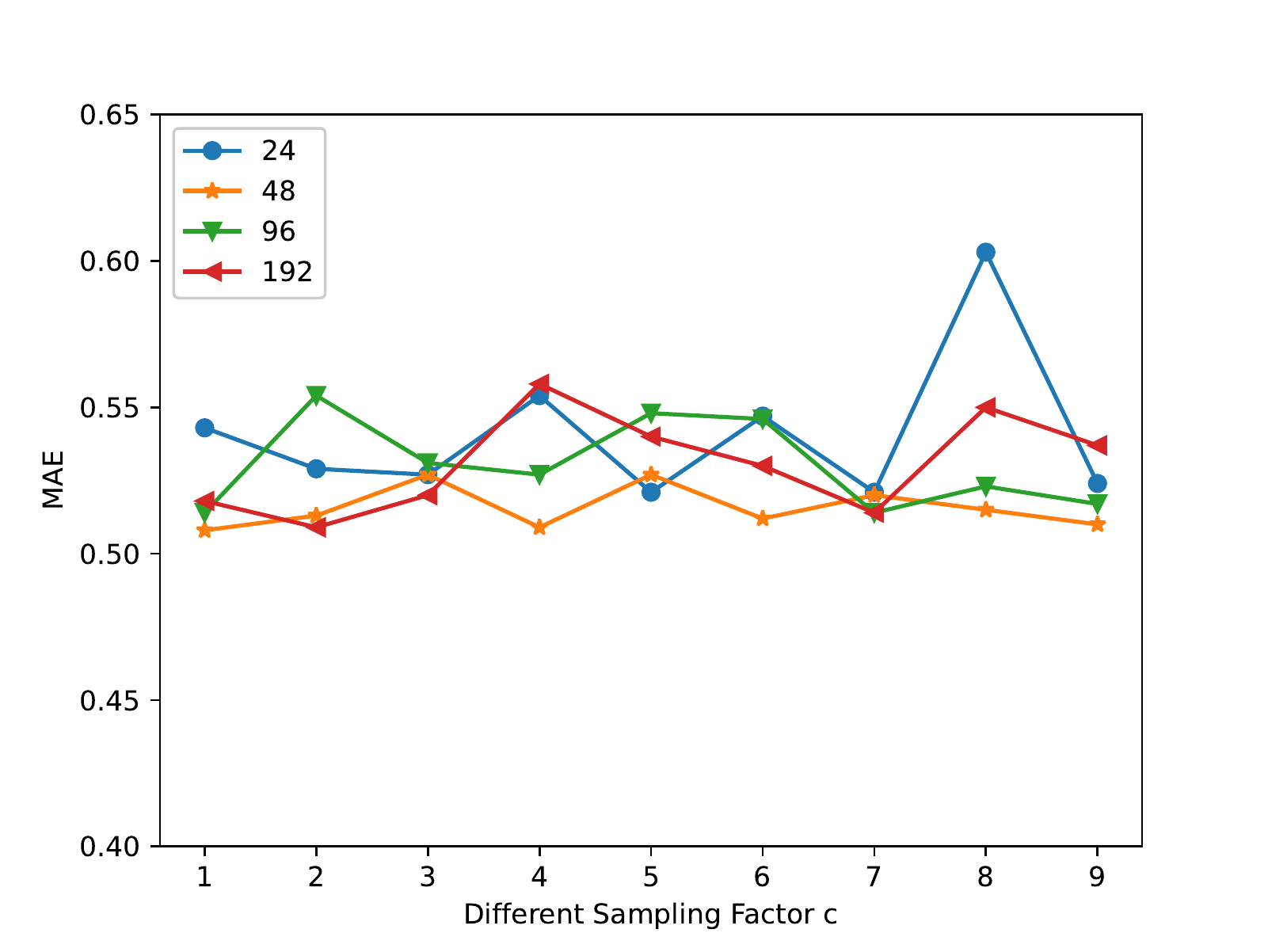}\vspace{1pt} 
		\end{minipage}
		\label{figc}
	}
	\subfigure[Sampling Factor $U$]{
		\begin{minipage}[t]{0.23\linewidth} 
			\includegraphics[width=\linewidth]{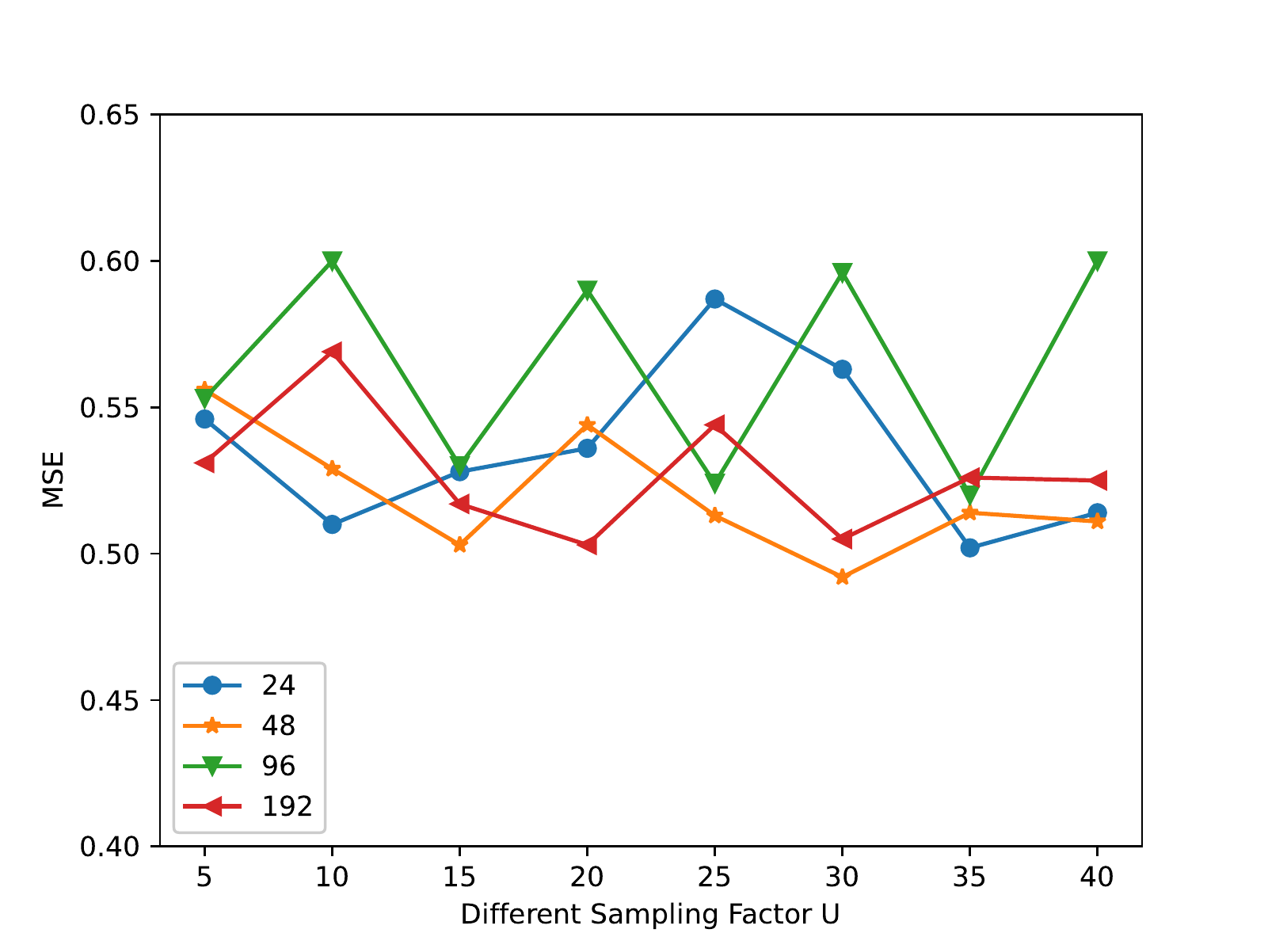}\vspace{1pt}
		\end{minipage}
		\begin{minipage}[t]{0.23\linewidth}
			\includegraphics[width=\linewidth]{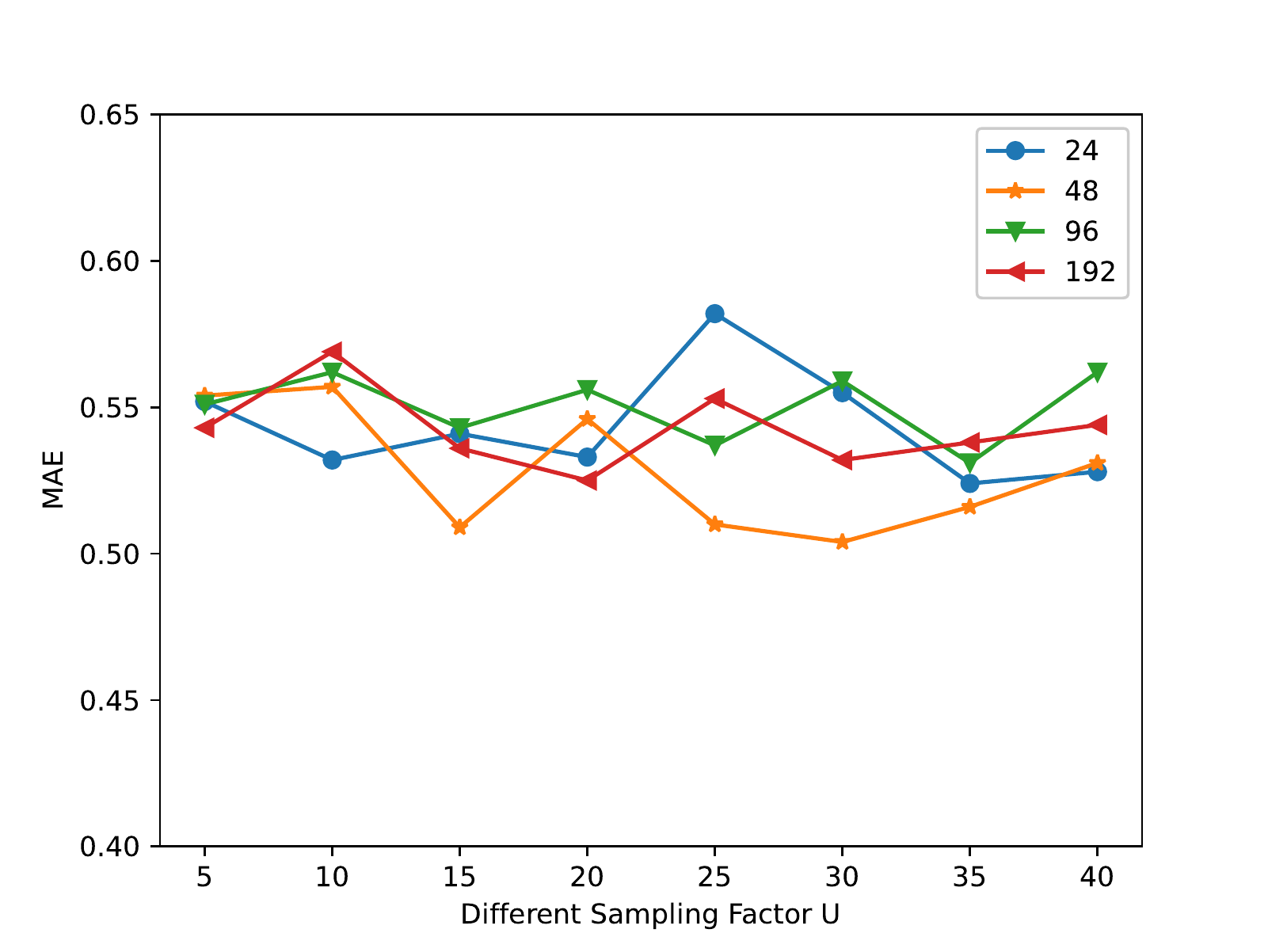}\vspace{1pt} 
		\end{minipage}
		\label{figu}
	}
	\caption{The parameter sensitivity of two components in Infomaxformer}
	\label{figps}
\end{figure*}

\subsection{Parameter Sensitivity}
We perform the sensitivity analysis of the proposed Infomaxformer model on ETTh$_1$.

\textbf{Input Length:}
As shown in the Table \ref{tablength}, we gradually extended the size of the input sequence $L_x$, i.e., \{336, 480, 720, 960, 1200, 1440\}, while keeping the predicted length $L_y$ unchanged, and the length $L_{label}$ of the label sequence is consistent with $L_y$. Our $L_y$ selected two values, 336 and 480. In Table \ref{tablength}, it can be seen that increasing the input length will lead to the decrease of MSE and MAE, because long input will bring repeated short-term patterns. However, as the input sequence increases, there may be more dependencies between the inputs, and the influence of noise in the input data will also increase. Some models can not effectively eliminate the influence of these noises and can not better grasp the dependence of long time-series, so MSE and MAE may increase. In this experiment, the batch size of Autoformer is set to 16, because too large batch size will directly lead to out-of-memory. In Table \ref{tabcomplexity}, we make statistics on the time complexity and memory usage of various models. It can be seen that there is still a certain gap between the memory usage of many model theories and the actual application. It seems that the Autoformer with memory usage of $\mathcal{O}(L\log L)$ is not better than the original Transformer with memory usage of $\mathcal{O}(L^2)$.

\textbf{Sampling Factor $c$:}
The sampling factor $c$ controls the information bandwidth of MEA in Equation (\ref{eq:method}). We start with small factors (=1) and gradually increase to large factors (=9). As can be seen in Figure \ref{figc}, the performance of our Infomaxformer does not change much, and it is not similar to the case that the performance of Informer is slightly improved with the change of sampling factor. This is because our initial sampling is $\sqrt{L}$, not $\log L$ in Informer \cite{zhou2021informer}, so enough dominant queries is selected to calculate the dot-product to prevent the loss of important features.

\textbf{Sampling Factor $U$:}
The sampling factor $U$ controls how many keys are selected to calculate the variance. Although the variance of data subject to the same distribution is the same, too few data samples will cause the calculated variance to be not the actual variance. It can be seen from Figure \ref{figu} that when $U$ is too low, the performance of the Infomaxformer model will indeed be affected. However, when $U$ gradually increases, the performance of the model tends to be stable.

\begin{table}[h]
	\centering
	\caption{Ablation study of the Infomaxformer}
	\resizebox{\linewidth}{!}{
		\begin{tabular}{cc|cccc}
			\toprule[1.0pt]
			\multicolumn{2}{c|}{Encoder's input}  & 720   & 960   & 1200  & 1440  \\
			\midrule[0.5pt]
			\multirow{2}{*}{Infomaxformer} & MSE & \textbf{0.566} & \textbf{0.588} & \textbf{0.633} & \textbf{0.573} \\
			& MAE & \textbf{0.579} & \textbf{0.585} & \textbf{0.617} & \textbf{0.585} \\
			\midrule[0.5pt]
			\multirow{2}{*}{Infomaxformer$^1$}              & MSE & 1.326 & 1.237 & 0.880 & 1.290 \\
			& MAE & 0.911 & 0.891 & 0.736 & 0.882 \\
			\midrule[0.5pt]
			\multirow{2}{*}{Infomaxformer$^2$}              & MSE &  0.906     & 0.681      & 0.839      & 0.831      \\
			& MAE &   0.763    &0.638       & 0.727      & 0.720      \\
			\midrule[0.5pt]
			\multirow{2}{*}{Infomaxformer$^3$}              & MSE & 1.074 & 1.018 & 0.964 & 1.121 \\
			& MAE & 0.822 & 0.823 & 0.789 & 0.846 \\
			\bottomrule[1.0pt] 
		\end{tabular}
	}

	\label{tabablation}
\end{table}
\subsection{Ablation Study}
We also performed some additional experiments for ablation analysis on ETTh$_1$. Infomaxformer$^1$ indicates that  Keys/Values Distilling is replaced with the original projection operation, Infomaxformer$^2$ indicates that MEA is replaced with the canonical self-attention mechanism, and Infomaxformer$^3$ indicates that we have not taken our proposed Time-Series Decomposition. In this experiment, we set the predicted length $L_y$ to 720 and select four ultra long input lengths. As shown in the Table \ref{tabablation}, without any part of the Infomaxformer model, the performance will be degraded. Only the complete Infomaxformer can achieve the best performance. The impact of MEA mechanism on the performance of Infomaxformer is not as obvious as the other two, because the mechanism focuses on sparing self-attention and reducing time complexity.
\begin{table}[h]
	\centering
	\caption{Comparative experiment of different Decomposition Block in Infomaxformer and Autoformer}
	\resizebox{\linewidth}{!}{
		\begin{tabular}{cc|cccc}
			\toprule[1.0pt]
			\multicolumn{2}{c|}{Predicition length}  & 96   & 192   & 384  & 720  \\
			\midrule[0.5pt]
			\multirow{2}{*}{TSD+MEA} & MSE & \textbf{0.554} & \textbf{0.496} & \textbf{0.567} & \textbf{0.551} \\
			& MAE & \textbf{0.540} & \textbf{0.513} & \textbf{0.555} & \textbf{0.559} \\
			\midrule[0.5pt]
			\multirow{2}{*}{SDB+MEA}              & MSE & 0.709 & 0.770 & 0.683 & 0.677 \\
			& MAE & 0.624 & 0.662 & 0.626 & 0.626 \\
			\midrule[0.5pt]
			\multirow{2}{*}{TSD+AC}              & MSE &  0.706     & 0.677      & -      & -      \\
			& MAE &   0.627    &0.605       & -      & -      \\
			\midrule[0.5pt]
			\multirow{2}{*}{SDB+AC}              & MSE & 0.623 & 0.699 & - & - \\
			& MAE & 0.577 & 0.619 & - & - \\
			\bottomrule[1.0pt] 
		\end{tabular}
	}

	\label{tababllst}
\end{table}

\textbf{Different Time-Series Decomposition:} In order to better compare the TSD proposed by us with the Series decomposition block (SDB) proposed by Autoformer \cite{wu2021autoformer}, we freely combined MEA, Auto-Correlation (AC) \cite{wu2021autoformer} with TSD, SDB. As shown in the Table \ref{tababllst}, we found an interesting phenomenon. No matter Infomaxformer (TSD+MEA) or Autoformer (SDB+AC), only the complete state model has the best performance. Our TSD is inferior to SDB in short sequence prediction ($L_y=96$), but superior to SDB in long sequence prediction ($L_y=192$). The sparsity of MEA results in the model being able to output longer sequences, but it is precisely because of this sparsity that the performance of MEA is inferior to AC when memory allows. However, the perfect combination of TSD and MEA proposed by us can output longer sequences and maintain better prediction performance.

\section{Conclusion}

In this paper, we studied the long time-series forecasting problem (LTFP) and proposed Infomaxformer to predict time-series. Specifically, we designed the Maximum Enterprise Self-attention mechanism and Keys/Values Distilling operation to deal with the challenges of quadratic time complexity and quadratic memory usage in vanilla Transformers. Finally, we reduce the time complexity and memory usage to $\mathcal{O}(L)$. In addition, the well-designed time-series decomposition and the perfect combination with Transformer architecture can effectively deal with the complex time-series patterns of time-series, thus alleviating the limitations of the traditional decomposition architecture. The experiments on real-word data have proved the effectiveness of Infomaxformer in improving the prediction ability of LTFP.



\bibliographystyle{ACM-Reference-Format}

\begin{thebibliography}{10}
	
	\bibitem{ba2016layer}
	{\sc J.~L. Ba, J.~R. Kiros, and G.~E. Hinton}, {\em Layer normalization}, arXiv
	preprint arXiv:1607.06450,  (2016).
	
	\bibitem{bahdanau2014neural}
	{\sc D.~Bahdanau, K.~Cho, and Y.~Bengio}, {\em Neural machine translation by
		jointly learning to align and translate}, arXiv preprint arXiv:1409.0473,
	(2014).
	
	\bibitem{bai2018empirical}
	{\sc S.~Bai, J.~Z. Kolter, and V.~Koltun}, {\em An empirical evaluation of
		generic convolutional and recurrent networks for sequence modeling}, arXiv
	preprint arXiv:1803.01271,  (2018).
	
	\bibitem{bao2021beit}
	{\sc H.~Bao, L.~Dong, and F.~Wei}, {\em Beit: Bert pre-training of image
		transformers}, arXiv preprint arXiv:2106.08254,  (2021).
	
	\bibitem{beltagy2020longformer}
	{\sc I.~Beltagy, M.~E. Peters, and A.~Cohan}, {\em Longformer: The
		long-document transformer}, arXiv preprint arXiv:2004.05150,  (2020).
	
	\bibitem{borovykh2017conditional}
	{\sc A.~Borovykh, S.~Bohte, and C.~W. Oosterlee}, {\em Conditional time series
		forecasting with convolutional neural networks}, arXiv preprint
	arXiv:1703.04691,  (2017).
	
	\bibitem{brown2020language}
	{\sc T.~Brown, B.~Mann, N.~Ryder, M.~Subbiah, J.~D. Kaplan, P.~Dhariwal,
		A.~Neelakantan, P.~Shyam, G.~Sastry, A.~Askell, et~al.}, {\em Language models
		are few-shot learners}, Advances in neural information processing systems, 33
	(2020), pp.~1877--1901.
	
	\bibitem{chen2020generative}
	{\sc M.~Chen, A.~Radford, R.~Child, J.~Wu, H.~Jun, D.~Luan, and I.~Sutskever},
	{\em Generative pretraining from pixels}, in International conference on
	machine learning, PMLR, 2020, pp.~1691--1703.
	
	\bibitem{child2019generating}
	{\sc R.~Child, S.~Gray, A.~Radford, and I.~Sutskever}, {\em Generating long
		sequences with sparse transformers}, arXiv preprint arXiv:1904.10509,
	(2019).
	
	\bibitem{cleveland1990stl}
	{\sc R.~B. Cleveland, W.~S. Cleveland, J.~E. McRae, and I.~Terpenning}, {\em
		Stl: A seasonal-trend decomposition}, J. Off. Stat, 6 (1990), pp.~3--73.
	
	\bibitem{cryer1986time}
	{\sc J.~D. Cryer}, {\em Time series analysis}, vol.~286, Springer, 1986.
	
	\bibitem{devlin2018bert}
	{\sc J.~Devlin, M.-W. Chang, K.~Lee, and K.~Toutanova}, {\em Bert: Pre-training
		of deep bidirectional transformers for language understanding}, arXiv
	preprint arXiv:1810.04805,  (2018).
	
	\bibitem{di2016artificial}
	{\sc L.~Di~Persio and O.~Honchar}, {\em Artificial neural networks
		architectures for stock price prediction: Comparisons and applications},
	International journal of circuits, systems and signal processing, 10 (2016),
	pp.~403--413.
	
	\bibitem{doersch2015unsupervised}
	{\sc C.~Doersch, A.~Gupta, and A.~A. Efros}, {\em Unsupervised visual
		representation learning by context prediction}, in Proceedings of the IEEE
	international conference on computer vision, 2015, pp.~1422--1430.
	
	\bibitem{2021An}
	{\sc A.~Dosovitskiy, L.~Beyer, A.~Kolesnikov, D.~Weissenborn, X.~Zhai,
		T.~Unterthiner, M.~Dehghani, M.~Minderer, G.~Heigold, and S.~a. Gelly}, {\em
		An image is worth 16x16 words: Transformers for image recognition at scale},
	in International Conference on Learning Representations, 2021.
	
	\bibitem{fan2019multi}
	{\sc C.~Fan, Y.~Zhang, Y.~Pan, X.~Li, C.~Zhang, R.~Yuan, D.~Wu, W.~Wang,
		J.~Pei, and H.~Huang}, {\em Multi-horizon time series forecasting with
		temporal attention learning}, in Proceedings of the 25th ACM SIGKDD
	International conference on knowledge discovery \& data mining, 2019,
	pp.~2527--2535.
	
	\bibitem{geng2022multimodal}
	{\sc X.~Geng, H.~Liu, L.~Lee, D.~Schuurams, S.~Levine, and P.~Abbeel}, {\em
		Multimodal masked autoencoders learn transferable representations}, arXiv
	preprint arXiv:2205.14204,  (2022).
	
	\bibitem{goyal2017accurate}
	{\sc P.~Goyal, P.~Doll{\'a}r, R.~Girshick, P.~Noordhuis, L.~Wesolowski,
		A.~Kyrola, A.~Tulloch, Y.~Jia, and K.~He}, {\em Accurate, large minibatch
		sgd: Training imagenet in 1 hour}, arXiv preprint arXiv:1706.02677,  (2017).
	
	\bibitem{guo2019star}
	{\sc Q.~Guo, X.~Qiu, P.~Liu, Y.~Shao, X.~Xue, and Z.~Zhang}, {\em
		Star-transformer}, in Proceedings of NAACL-HLT, 2019, pp.~1315--1325.
	
	\bibitem{he2022masked}
	{\sc K.~He, X.~Chen, S.~Xie, Y.~Li, P.~Doll{\'a}r, and R.~Girshick}, {\em
		Masked autoencoders are scalable vision learners}, in Proceedings of the
	IEEE/CVF Conference on Computer Vision and Pattern Recognition, 2022,
	pp.~16000--16009.
	
	\bibitem{he2016deep}
	{\sc K.~He, X.~Zhang, S.~Ren, and J.~Sun}, {\em Deep residual learning for
		image recognition}, in Proceedings of the IEEE conference on computer vision
	and pattern recognition, 2016, pp.~770--778.
	
	\bibitem{hinton1993autoencoders}
	{\sc G.~E. Hinton and R.~Zemel}, {\em Autoencoders, minimum description length
		and helmholtz free energy}, Advances in neural information processing
	systems, 6 (1993).
	
	\bibitem{hochreiter2001gradient}
	{\sc S.~Hochreiter, Y.~Bengio, P.~Frasconi, J.~Schmidhuber, et~al.}, {\em
		Gradient flow in recurrent nets: the difficulty of learning long-term
		dependencies}, 2001.
	
	\bibitem{hochreiter1997long}
	{\sc S.~Hochreiter and J.~Schmidhuber}, {\em Long short-term memory}, Neural
	computation, 9 (1997), pp.~1735--1780.
	
	\bibitem{hyndman2018forecasting}
	{\sc R.~J. Hyndman and G.~Athanasopoulos}, {\em Forecasting: principles and
		practice}, OTexts, 2018.
	
	\bibitem{loshchilov2016sgdr}
	{\sc {I. Loshchilov and F. Hutter}}, {\em Sgdr: Stochastic gradient descent
		with warm restarts}, in ICLR, 2017.
	
	\bibitem{jaynes1957information}
	{\sc E.~T. Jaynes}, {\em Information theory and statistical mechanics},
	Physical review, 106 (1957), p.~620.
	
	\bibitem{jiang2022accurate}
	{\sc H.~Jiang, M.~Cui, D.~W.~K. Ng, and L.~Dai}, {\em Accurate channel
		prediction based on transformer: Making mobility negligible}, IEEE Journal on
	Selected Areas in Communications,  (2022).
	
	\bibitem{kingma2014adam}
	{\sc D.~P. Kingma and J.~Ba}, {\em Adam: {A} method for stochastic
		optimization}, in International Conference on Learning Representations, 2015.
	
	\bibitem{kitaev2019reformer}
	{\sc N.~Kitaev, L.~Kaiser, and A.~Levskaya}, {\em Reformer: The efficient
		transformer}, in ICLR, 2019.
	
	\bibitem{krizhevsky2012imagenet}
	{\sc A.~Krizhevsky, I.~Sutskever, and G.~E. Hinton}, {\em Imagenet
		classification with deep convolutional neural networks}, Advances in neural
	information processing systems, 25 (2012).
	
	\bibitem{lai2018modeling}
	{\sc G.~Lai, W.-C. Chang, Y.~Yang, and H.~Liu}, {\em Modeling long-and
		short-term temporal patterns with deep neural networks}, in The 41st
	international ACM SIGIR conference on research \& development in information
	retrieval, 2018, pp.~95--104.
	
	\bibitem{li2019enhancing}
	{\sc S.~Li, X.~Jin, Y.~Xuan, X.~Zhou, W.~Chen, Y.-X. Wang, and X.~Yan}, {\em
		Enhancing the locality and breaking the memory bottleneck of transformer on
		time series forecasting}, Advances in neural information processing systems,
	32 (2019).
	
	\bibitem{li2022uniform}
	{\sc X.~Li, W.~Wang, L.~Yang, and J.~Yang}, {\em Uniform masking: Enabling mae
		pre-training for pyramid-based vision transformers with locality}, arXiv
	preprint arXiv:2205.10063,  (2022).
	
	\bibitem{lim2021temporal}
	{\sc B.~Lim, S.~{\"O}. Ar{\i}k, N.~Loeff, and T.~Pfister}, {\em Temporal fusion
		transformers for interpretable multi-horizon time series forecasting},
	International Journal of Forecasting, 37 (2021), pp.~1748--1764.
	
	\bibitem{lim2021time}
	{\sc B.~Lim and S.~Zohren}, {\em Time-series forecasting with deep learning: a
		survey}, Philosophical Transactions of the Royal Society A, 379 (2021),
	p.~20200209.
	
	\bibitem{lim2020recurrent}
	{\sc B.~Lim, S.~Zohren, and S.~Roberts}, {\em Recurrent neural filters:
		Learning independent bayesian filtering steps for time series prediction}, in
	2020 International Joint Conference on Neural Networks (IJCNN), IEEE, 2020,
	pp.~1--8.
	
	\bibitem{lin2021survey}
	{\sc T.~Lin, Y.~Wang, X.~Liu, and X.~Qiu}, {\em A survey of transformers},
	arXiv preprint arXiv:2106.04554,  (2021).
	
	\bibitem{liu2018learning}
	{\sc L.~Liu, J.~Shen, M.~Zhang, Z.~Wang, and J.~Tang}, {\em Learning the joint
		representation of heterogeneous temporal events for clinical endpoint
		prediction}, in Proceedings of the AAAI Conference on Artificial
	Intelligence, vol.~32, 2018.
	
	\bibitem{liu2021pyraformer}
	{\sc S.~Liu, H.~Yu, C.~Liao, J.~Li, W.~Lin, A.~X. Liu, and S.~Dustdar}, {\em
		Pyraformer: Low-complexity pyramidal attention for long-range time series
		modeling and forecasting}, in International Conference on Learning
	Representations, 2021.
	
	\bibitem{liu2021swin}
	{\sc Z.~Liu, Y.~Lin, Y.~Cao, H.~Hu, Y.~Wei, Z.~Zhang, S.~Lin, and B.~Guo}, {\em
		Swin transformer: Hierarchical vision transformer using shifted windows}, in
	Proceedings of the IEEE/CVF International Conference on Computer Vision,
	2021, pp.~10012--10022.
	
	\bibitem{loshchilov2017decoupled}
	{\sc I.~Loshchilov and F.~Hutter}, {\em Decoupled weight decay regularization},
	in ICLR, 2019.
	
	\bibitem{1937Jensen}
	{\sc E.~J. Mcshane}, {\em Jensen's inequality}, Bulletin of the American
	Mathematical Society, 43 (1937), pp.~521--527.
	
	\bibitem{noroozi2016unsupervised}
	{\sc M.~Noroozi and P.~Favaro}, {\em Unsupervised learning of visual
		representations by solving jigsaw puzzles}, in European conference on
	computer vision, Springer, 2016, pp.~69--84.
	
	\bibitem{parmar2018image}
	{\sc N.~Parmar, A.~Vaswani, J.~Uszkoreit, L.~Kaiser, N.~Shazeer, A.~Ku, and
		D.~Tran}, {\em Image transformer}, in International conference on machine
	learning, PMLR, 2018, pp.~4055--4064.
	
	\bibitem{paszke2019pytorch}
	{\sc A.~Paszke, S.~Gross, F.~Massa, A.~Lerer, J.~Bradbury, G.~Chanan,
		T.~Killeen, Z.~Lin, N.~Gimelshein, L.~Antiga, et~al.}, {\em Pytorch: An
		imperative style, high-performance deep learning library}, Advances in neural
	information processing systems, 32 (2019).
	
	\bibitem{radford2018improving}
	{\sc A.~Radford, K.~Narasimhan, T.~Salimans, I.~Sutskever, et~al.}, {\em
		Improving language understanding by generative pre-training},  (2018).
	
	\bibitem{radford2019language}
	{\sc A.~Radford, J.~Wu, R.~Child, D.~Luan, D.~Amodei, I.~Sutskever, et~al.},
	{\em Language models are unsupervised multitask learners}, OpenAI blog, 1
	(2019), p.~9.
	
	\bibitem{rangapuram2018deep}
	{\sc S.~S. Rangapuram, M.~W. Seeger, J.~Gasthaus, L.~Stella, Y.~Wang, and
		T.~Januschowski}, {\em Deep state space models for time series forecasting},
	Advances in neural information processing systems, 31 (2018).
	
	\bibitem{roy2021efficient}
	{\sc A.~Roy, M.~Saffar, A.~Vaswani, and D.~Grangier}, {\em Efficient
		content-based sparse attention with routing transformers}, Transactions of
	the Association for Computational Linguistics, 9 (2021), pp.~53--68.
	
	\bibitem{salinas2020deepar}
	{\sc D.~Salinas, V.~Flunkert, J.~Gasthaus, and T.~Januschowski}, {\em Deepar:
		Probabilistic forecasting with autoregressive recurrent networks},
	International Journal of Forecasting, 36 (2020), pp.~1181--1191.
	
	\bibitem{shannon1948mathematical}
	{\sc C.~E. Shannon}, {\em A mathematical theory of communication}, The Bell
	system technical journal, 27 (1948), pp.~379--423.
	
	\bibitem{tang2022features}
	{\sc P.~Tang and X.~Zhang}, {\em Features fusion framework for multimodal
		irregular time-series events}, in Pacific Rim International Conference on
	Artificial Intelligence, Springer, 2022, pp.~366--379.
	
	\bibitem{tang2022mtsmae}
	\leavevmode\vrule height 2pt depth -1.6pt width 23pt, {\em Mtsmae: Masked
		autoencoders for multivariate time-series forecasting}, arXiv preprint
	arXiv:2210.02199,  (2022).
	
	\bibitem{tsai2019transformer}
	{\sc Y.-H.~H. Tsai, S.~Bai, M.~Yamada, L.-P. Morency, and R.~Salakhutdinov},
	{\em Transformer dissection: An unified understanding for transformer’s
		attention via the lens of kernel}, in Proceedings of the 2019 Conference on
	Empirical Methods in Natural Language Processing and the 9th International
	Joint Conference on Natural Language Processing (EMNLP-IJCNLP), 2019,
	pp.~4344--4353.
	
	\bibitem{vaswani2017attention}
	{\sc A.~Vaswani, N.~Shazeer, N.~Parmar, J.~Uszkoreit, L.~Jones, A.~N. Gomez,
		{\L}.~Kaiser, and I.~Polosukhin}, {\em Attention is all you need}, Advances
	in neural information processing systems, 30 (2017).
	
	\bibitem{wang2015unsupervised}
	{\sc X.~Wang and A.~Gupta}, {\em Unsupervised learning of visual
		representations using videos}, in Proceedings of the IEEE international
	conference on computer vision, 2015, pp.~2794--2802.
	
	\bibitem{wu2021autoformer}
	{\sc H.~Wu, J.~Xu, J.~Wang, and M.~Long}, {\em Autoformer: Decomposition
		transformers with auto-correlation for long-term series forecasting},
	Advances in Neural Information Processing Systems, 34 (2021),
	pp.~22419--22430.
	
	\bibitem{wu2020adversarial}
	{\sc S.~Wu, X.~Xiao, Q.~Ding, P.~Zhao, Y.~Wei, and J.~Huang}, {\em Adversarial
		sparse transformer for time series forecasting}, Advances in neural
	information processing systems, 33 (2020), pp.~17105--17115.
	
	\bibitem{xie2022simmim}
	{\sc Z.~Xie, Z.~Zhang, Y.~Cao, Y.~Lin, J.~Bao, Z.~Yao, Q.~Dai, and H.~Hu}, {\em
		Simmim: A simple framework for masked image modeling}, in Proceedings of the
	IEEE/CVF Conference on Computer Vision and Pattern Recognition, 2022,
	pp.~9653--9663.
	
	\bibitem{ye2019bp}
	{\sc Z.~Ye, Q.~Guo, Q.~Gan, X.~Qiu, and Z.~Zhang}, {\em Bp-transformer:
		Modelling long-range context via binary partitioning}, arXiv preprint
	arXiv:1911.04070,  (2019).
	
	\bibitem{yin2016multivariate}
	{\sc Y.~Yin and P.~Shang}, {\em Multivariate multiscale sample entropy of
		traffic time series}, Nonlinear Dynamics, 86 (2016), pp.~479--488.
	
	\bibitem{yu2021glance}
	{\sc Q.~Yu, Y.~Xia, Y.~Bai, Y.~Lu, A.~L. Yuille, and W.~Shen}, {\em
		Glance-and-gaze vision transformer}, Advances in Neural Information
	Processing Systems, 34 (2021), pp.~12992--13003.
	
	\bibitem{zaheer2020big}
	{\sc M.~Zaheer, G.~Guruganesh, K.~A. Dubey, J.~Ainslie, C.~Alberti, S.~Ontanon,
		P.~Pham, A.~Ravula, Q.~Wang, L.~Yang, et~al.}, {\em Big bird: Transformers
		for longer sequences}, Advances in Neural Information Processing Systems, 33
	(2020), pp.~17283--17297.
	
	\bibitem{zhou2021informer}
	{\sc H.~Zhou, S.~Zhang, J.~Peng, S.~Zhang, J.~Li, H.~Xiong, and W.~Zhang}, {\em
		Informer: Beyond efficient transformer for long sequence time-series
		forecasting}, in Proceedings of the AAAI Conference on Artificial
	Intelligence, vol.~35, 2021, pp.~11106--11115.
	
\end{thebibliography}


\end{document}